\documentclass[sigconf]{acmart}

\makeatletter
\def\@ACM@checkaffil{
    \if@ACM@instpresent\else
    \ClassWarningNoLine{\@classname}{No institution present for an affiliation}%
    \fi
    \if@ACM@citypresent\else
    \ClassWarningNoLine{\@classname}{No city present for an affiliation}%
    \fi
    \if@ACM@countrypresent\else
        \ClassWarningNoLine{\@classname}{No country present for an affiliation}%
    \fi
}
\makeatother

\usepackage{balance}
\usepackage{float}
\usepackage{graphicx}
\usepackage{subfigure}
\usepackage{diagbox}

\usepackage[ruled,linesnumbered]{algorithm2e}
\usepackage{subfigure}

\newcommand{\Rmnum}[1]{\expandafter\@slowromancap\romannumeral #1@}

\newcommand{\eg}{\emph{e.g.},\xspace}

\newcommand{\etc}{\emph{etc.}\xspace}

\DeclareMathAlphabet\mathbfcal{OMS}{cmsy}{b}{n}

\newcommand{\eat}[1]{}
\ifodd 1

\else

\fi

\AtBeginDocument{%
  \providecommand\BibTeX{{%
    \normalfont B\kern-0.5em{\scshape i\kern-0.25em b}\kern-0.8em\TeX}}}

\copyrightyear{2023}
\acmYear{2023}
\setcopyright{acmlicensed}
\acmConference[KDD '23] {Proceedings of the 29th ACM SIGKDD Conference on Knowledge Discovery and Data Mining}{August 6--10, 2023}{Long Beach, CA, USA.}
\acmBooktitle{Proceedings of the 29th ACM SIGKDD Conference on Knowledge Discovery and Data Mining (KDD '23), August 6--10, 2023, Long Beach, CA, USA}
\acmPrice{15.00}
\acmISBN{979-8-4007-0103-0/23/08}
\acmDOI{10.1145/3580305.3599492}

\begin{document}
\title{Robust Spatiotemporal Traffic Forecasting with Reinforced Dynamic Adversarial Training}
\thanks{$^*$ Corresponding author.}
\author{Fan Liu}
\affiliation{\institution{Artificial Intelligence Thrust, The Hong Kong University of Science and Technology (Guangzhou)}}
\email{fliu236@connect.hkust-gz.edu.cn}

\author{Weijia Zhang}
\affiliation{\institution{Artificial Intelligence Thrust, The Hong Kong University of Science and Technology (Guangzhou)}}
\email{wzhang411@connect.hkust-gz.edu.cn}

\author{Hao Liu$^{*}$}
\affiliation{\institution{Artificial Intelligence Thrust, The Hong Kong University of Science and Technology (Guangzhou)} \institution{Department of Computer Science and Engineering, The Hong Kong University of Science and Technology}}
\email{liuh@ust.hk}

\renewcommand{\shortauthors}{Fan Liu, Weijia Zhang, \& Hao Liu}

\begin{abstract}
Machine learning-based forecasting models are commonly used in Intelligent Transportation Systems (ITS) to predict traffic patterns and provide city-wide services. However, most of the existing models are susceptible to adversarial attacks, which can lead to inaccurate predictions and negative consequences such as congestion and delays.  Therefore, improving the adversarial robustness of these models is crucial for ITS. In this paper, we propose a novel framework for incorporating adversarial training into spatiotemporal traffic forecasting tasks. We demonstrate that traditional adversarial training methods designated for static domains cannot be directly applied to traffic forecasting tasks, as they fail to effectively defend against dynamic adversarial attacks. Then, we propose a reinforcement learning-based method to learn the optimal node selection strategy for adversarial examples, which simultaneously strengthens the dynamic attack defense capability and reduces the model overfitting. Additionally, we introduce a self-knowledge distillation regularization module to overcome the "forgetting issue" caused by continuously changing adversarial nodes during training.   We evaluate our approach on two real-world traffic datasets and demonstrate its superiority over other baselines. Our method effectively enhances the adversarial robustness of spatiotemporal traffic forecasting models. The source code for our framework is available at \url{https://github.com/usail-hkust/RDAT}.
\end{abstract}




\maketitle

\section{Introduction}
\begin{figure}[t]
    \centering
    \scalebox{0.26}{\includegraphics{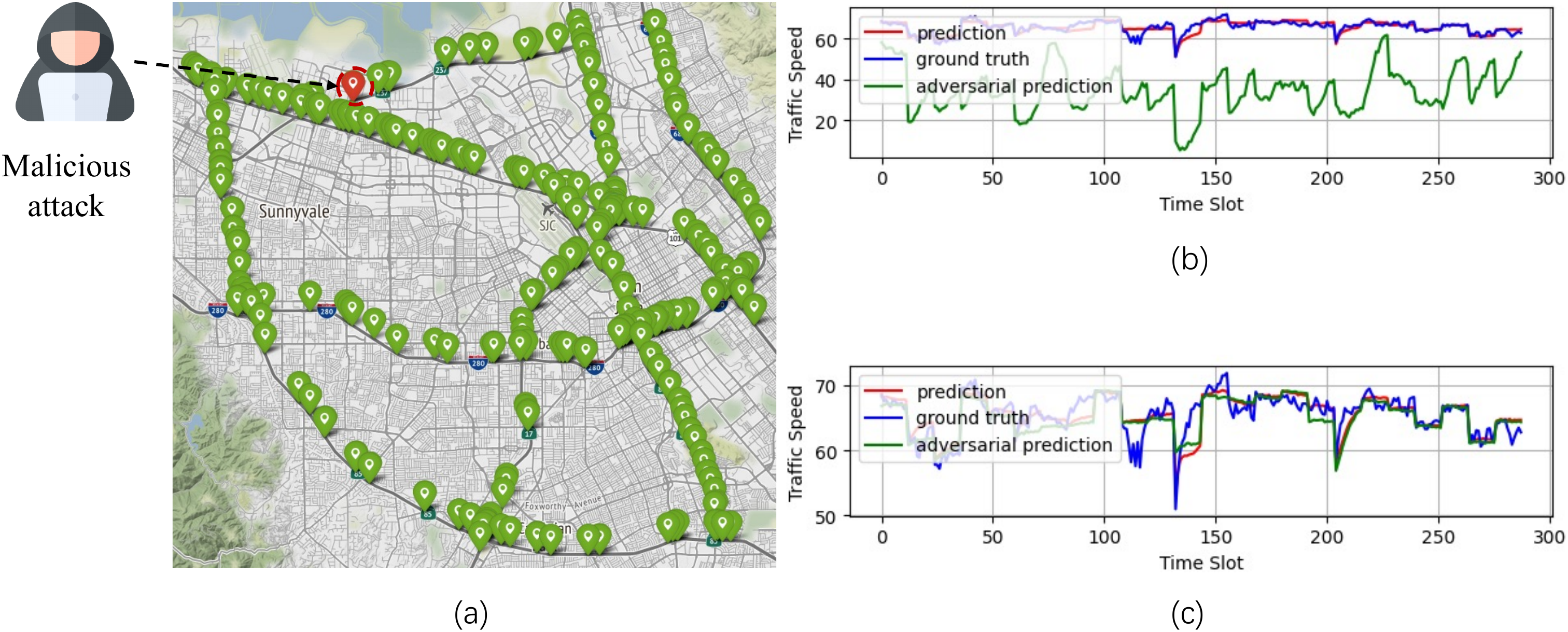}}    \caption{  An example of adversarial attack and defense on spatiotemporal traffic forecasting model using the PeMS-Bay dataset. (a) The adversarial attack injects malicious perturbations into the geo-distributed data sources. (b) As a result, the forecasting model's performance degrades and produces biased predictions under the adversarial attack. (c) The defense against adversarial attacks is achieved through adversarial training, resulting in improved model performance and accurate traffic predictions.}
    \label{fig:malicious}
    \setlength{\belowcaptionskip}{-0.5cm}
\end{figure}

Machine learning-based forecasting models are widely used to accurately and promptly predict traffic patterns for providing city-wide services in intelligent traffic systems (ITS)~\cite{STGCN, liu2021community, zhang2022multi}. However, these models can be fooled by carefully crafted perturbations,  leading to inaccurate traffic conditions predictions~\cite{liu2022practical}. For example,  a small perturbation to a traffic flow map can stimulate a machine learning model to predict a traffic jam where there was none, leading to unnecessary congestion and delays. Figure~\ref{fig:malicious} demonstrates the impact of an adversarial attack on the spatiotemporal forecasting model, resulting in significant bias in predictions. Fortunately, recent studies identify that incorporating a defense strategy can effectively improve the adversarial robustness of machine learning models~\cite{liu2022practical}. Therefore, there is a pressing need to investigate suitable defense strategies to stabilize the  spatiotemporal forecasting models, particularly for ITS.

Adversarial training is a technique that has been shown to enhance the robustness of deep neural networks (DNNs) against adversarial attacks, particularly in static domains such as image~\cite{PangZJLW20, ChenG20, ShuklaSWK21} and graph~\cite{ZhuZ0019, ZugnerG20, WangJCG21, WDWT21} classification. This is achieved by incorporating adversarial examples, generated through adversarial attacks, into the training process. The adversarial training is formulated as a min-max optimization problem, where the inner maximization step generates adversarial examples to explore worst-case scenarios within the adversarial perturbation space. These small, yet perceptible, perturbations are designed to cause the model to make incorrect predictions. In the outer minimization step, the model is exposed to both the original input data and the adversarial examples, which are used to update the model and improve its overall robustness against such perturbations. Despite its efficacy in static domains~\cite{WangHBSML020}, adversarial training for spatiotemporal traffic forecasting remains under-explored in dynamic fields.

\eat{
Recent research has demonstrated that adversarial training can effectively stabilize the adversarial robustness of deep neural networks (DNNs), especially in static domains ( \eg image and graph classification tasks \etc). It involves incorporating adversarial examples generated through adversarial attacks into the training process, thus enhancing the DNN's ability to defend against adversarial attacks.  The adversarial training can be formulated as a min-max optimization problem, where the inner maximization process involves generating adversarial examples that explore the worst-case scenarios within the adversarial perturbation space. The adversarial examples are designed to be small perturbations to the input data that are imperceptible to humans but that can cause the model to make incorrect predictions. In the outer minimization step of adversarial training, the deep learning model is exposed to both the original input data and adversarial examples. These examples are used to update the model, enabling it to better handle such perturbations and improve its overall robustness. The ultimate goal of adversarial training is to enhance the model's adversarial robustness to adversarial attacks. However, adversarial training for spatiotemporal traffic forecasting  is less explored in dynamic fields. }

\begin{figure}[t]
\centering
\subfigure[]{
\begin{minipage}[]{0.5\linewidth}
\centering
\includegraphics[width=1.4in]{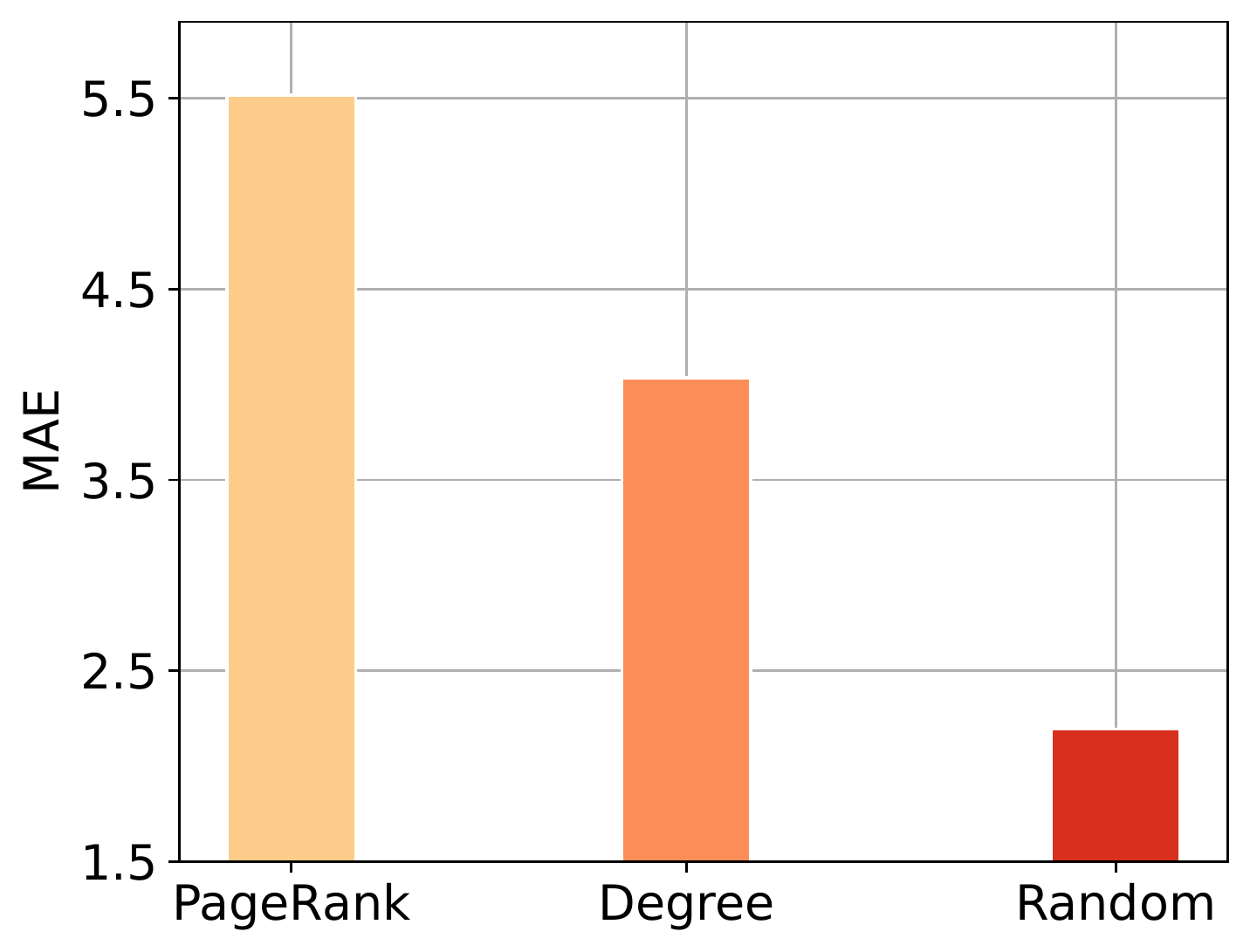}
\end{minipage}%
}%
\subfigure[]{
\begin{minipage}[]{0.5\linewidth}
\centering
\includegraphics[width=1.4in]{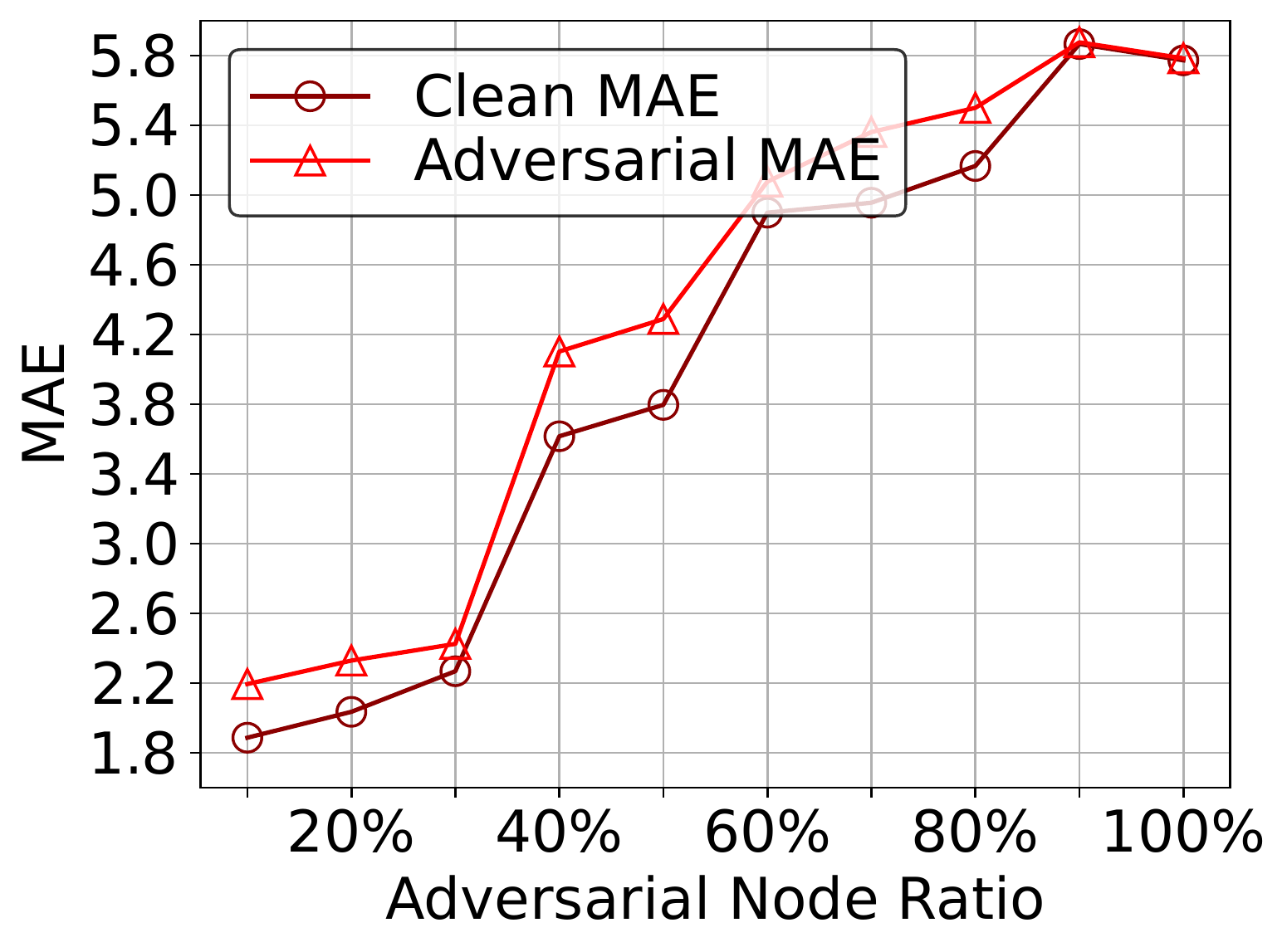}
\end{minipage}%
}%
\centering
\setlength{\belowcaptionskip}{-0.7cm}
\caption{ (a) Comparison of model performance under static (PageRank and Degree) and dynamic (Random) defense strategies in adversarial attacks. The graph shows the results of the traffic forecasting model under an adversarial attack, with defense using static PageRank and degree-based sampling strategies (left two bars) and dynamic random sampling strategy (right bar). It is evident that the dynamic random sampling strategy results in better performance than the static defense strategies.
(b) Comparison of model performance under varying levels of adversarial node ratio while maintaining a consistent attack strength. The X-axis shows the proportion of adversarial nodes, and the Y-axis shows the model's performance.  A lower proportion of adversarial nodes results in better model performance compared to a higher proportion of adversarial nodes.  (Clean MAE refers to the model's performance when evaluated on clean examples, whereas Adversarial MAE indicates the model's performance when evaluated on adversarial examples.)}\label{fig:Dymamic_and_static_dif_proportion}
\end{figure}

In this paper, we reveal the limitations of traditional adversarial training methods in defending against dynamic adversarial attacks in spatiotemporal traffic forecasting tasks. We demonstrate that static strategies for selecting and generating adversarial examples, such as using degree and PageRank, fail to effectively defend against these attacks as shown in Figure~\ref{fig:Dymamic_and_static_dif_proportion} (a).  First, we identify the static approach can not  provide adequate defense against spatiotemporal adversarial attacks.  Furthermore, we show that generating adversarial examples for all geo-distributed data sources also fails to effectively defend against dynamic attacks, as a higher proportion of adversarial nodes may lead to overfitting and reduces model performance as shown in Figure~\ref{fig:Dymamic_and_static_dif_proportion} (b).  Counterintuitively, a lower proportion of adversarial nodes results in better model performance compared to a higher proportion of adversarial nodes.  Additionally, we highlight the issue of instability that arises when adversarial nodes change continuously during the training process, resulting in the "forgetting issue" where the model lacks robustness against stronger attack strengths. Such observations highlight highlights the need for an effective and robust approach to adversarial training in spatiotemporal traffic forecasting tasks.

To overcome the aforementioned limitations, we propose a novel framework for incorporating adversarial training into traffic forecasting tasks. Our approach involves dynamically selecting a subset of nodes as adversarial examples, which not only reduces overfitting but also improves defense capability against dynamic adversarial attacks. However, the task of selecting this subset from the total set of nodes is a computationally challenging problem, known to be NP-hard. To address this issue, we propose a reinforcement learning-based method to learn the optimal node selection strategy. Specifically, we model the node selection problem as a combinatorial optimization problem and use a policy-based network to learn the node selection strategy that maximizes the inner loss. In detail, we design a spatiotemporal attention-based policy network to model spatiotemporal geo-distributed data. To evaluate the solutions generated by the policy network, we propose a balanced strategy for the reward function, providing stable and efficient feedback to the policy network and mitigating the issue of  inner loss decreasing during training. The final pre-trained policy network can be used as a node selector. To overcome the forgetting issue, we also introduce a new self-knowledge distillation  regularization module for adversarial training, where the current model is trained using knowledge distilled from the previous model's experience with adversarial attacks.

Our contributions can be summarized as follows.  
\textbf{1).}  To our knowledge, we are the first to defend against adversarial attacks for spatiotemporal traffic forecasting by systematically analyzing how to apply adversarial training to traffic forecasting models.
\textbf{2).}  We propose a novel framework for improving the adversarial robustness of spatiotemporal traffic forecasting. This includes modeling the node selection problem as a combinatorial optimization problem and using a reinforcement learning-based method to learn the optimal node selection strategy. Furthermore, we incorporate self-knowledge distillation as a new training technique to tackle the challenge of continuously evolving adversarial nodes, thus avoiding the "forgetting issue.  \textbf{3).} We conduct extensive experiments on two real-world datasets and demonstrate the effectiveness of our framework and its individual components.

\section{Notation and Preliminaries}
We first provide an overview of nations, and then delve into the topics of spatiotemporal traffic  forecasting, adversarial training, and the threat model.

The traffic network can be represented by the graph $\mathcal{G} = (\mathcal{V}, \mathcal{E})$, where $\mathcal{V}$ is a set of $n$ nodes (such as traffic sensors, road stretches, highway segments, \etc) and $\mathcal{E}$ is a set of edges. We denote the adjacency matrix $\mathcal{A}$ to represent the  traffic network  $\mathcal{G} = (\mathcal{V}, \mathcal{E})$.
Furthermore, we use the $\mathcal{X}_{t}= \{ \mathbf{x}_{1,t},\mathbf{x}_{2,t},\mathbf{x}_{3,t},\cdots, \mathbf{x}_{n,t} \}$ geo-distributed data features, where $\mathbf{x}_{i,t}  $ represents the traffic conditions (\eg  traffic speed, the volume of vehicles, and traffic flow \etc) and context information (\eg point of interest, road closures, construction, and accidents \etc ) for node ${v}_{i}$.
\subsection{Spatiotemporal Traffic Forecasting}
Given the historical $\tau$ traffic situations, the spatiotemporal traffic forecasting models aim to predict the next $T$ traffic situations,

\begin{equation}
    \mathbfcal{\hat{Y}}_{t+1:t+T} = \mathcal{F}_{\theta}(\mathbfcal{X}_{t-\tau:t} ;\mathbfcal{A}),
\end{equation}
where $\mathbfcal{\hat{Y}}_{t+1:t+T} = \{ \mathcal{\hat{Y}}_{t+1}, \cdots,\mathcal{\hat{Y}}_{t+T}\} $ is the predicted traffic conditions from time slot $t$ to $t+T$. $\mathcal{F}_{\theta}(\cdot)$ is the spatiotemporal traffic forecasting models with model parameter $\theta$. $\mathbfcal{Y}_{t+1:t+T} = \{ \mathcal{Y}_{t+1}, \cdots,\mathcal{Y}_{t+T}\} $ is the ground truth for spatiotemporal features  $\mathbfcal{X}_{t-\tau:t}  = \{\mathcal{X}_{t-1}, \mathcal{X}_{t}, \cdots ,\\ \mathcal{X}_{t}\}$.

\begin{figure*}[t]
    \centering
    \scalebox{0.30}{\includegraphics{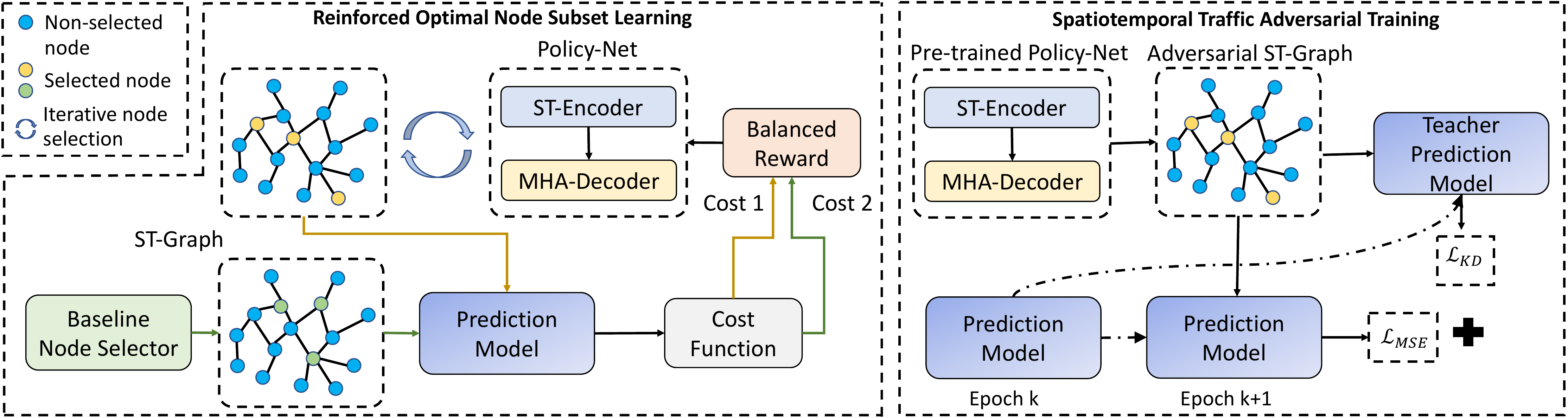}}
    \setlength{\belowcaptionskip}{-0.5cm}
    \caption{The framework overview of RDAT. }
    \label{fig:method}
\end{figure*}

\subsection{Adversarial Training}
Adversarial training involves using adversarial examples, which are generated through adversarial attacks, in the training process to improve the model's robustness.
The adversarial training can be formulated as a min-max optimization problem, 
\begin{equation}
\min_{\theta}\max_{x' \in \mathcal{B}(x, \epsilon) }\mathcal{L}(f_{\theta }(x'),y)
\end{equation}
where $\theta$ represents the model parameters, $x'$ denotes the adversarial example, and $\mathcal{B}(x, \epsilon)=\{ x + \delta \mid \| \delta \|_{p} \le \epsilon \}$ represents the set of allowed adversarial example set with a maximum perturbation budget of $\epsilon$, where $\delta$ denotes the adversarial perturbation. The  $f_{\theta}(\cdot)$ represents the deep learning model, and $y$ represents the ground truth.  \eat{The inner maximization generates the adversarial examples, and the outer minimization updates the model using these examples to improve its robustness.}

\subsection{Threat Model}
We follow the taxonomy of adversarial attacks on spatiotemporal traffic forecasting models as presented in \cite{liu2022practical}.
\textbf{Attacker's goal.} The goal of the attacker is to create adversarial traffic states that will cause spatiotemporal forecasting models to derive biased predictions.
\textbf{Evasion attack.} The attack is launched during the inference stage after the model has already been trained. 
\textbf{Attacker's capability.} Our focus is on spatiotemporal feature-level attacks, in which the attacker can alter the spatiotemporal features by injecting adversarial perturbations into geo-distributed data sources. We do not focus on graph structure-level attacks because spatiotemporal feature-level attacks can lead to higher attack success rates. Note that the attacker does not have the ability to manipulate the model's architecture or parameters during the attack.
\textbf{Attack strength.} The attacker can inject adversarial perturbations into $\lambda \%$ of the geo-distributed data sources. As the proportion of hacked geo-distributed data sources increases, the attack becomes stronger and more intense.
\eat{\textbf{Attacker's knowledge.} Based on the level of knowledge possessed by the adversary, the attack can be divided into white-box attack and black-box attack. In the white-box attack setting, the attacker has full knowledge of the model parameters, model type, and training data, and also has access to the output of the target model. In the black-box attack setting, the attacker has no knowledge of the model parameters but has access to the training data and can query the output of the target model.}

\textbf{Problem definition}: The goal of this research is to develop an adversarial robust spatiotemporal traffic forecasting model, denoted as $\mathcal{F}_{\theta}(\cdot)$, that is capable of defending against adversarial attacks on geo-distributed data sources.

\section{Methodology}
This section presents an experimental investigation of the use of adversarial training for spatiotemporal traffic forecasting. Our proposed framework, outlined in detail in Section~\ref{sec:sub-ATF}, centers on the dynamic selection of a subset of nodes as adversarial nodes.  In Section~\ref{sec:sub-RONSL}, we mathematically model the selection of the optimal subset of nodes as a combinatorial optimization problem and introduce a spatiotemporal attention-based representation module to improve the learning of node representations and aid in policy learning. To address the issue of instability, Section~\ref{sec:sub-ATR} introduces a self-distillation regularization term to prevent forgetting. 

\subsection{Framework Overview}\label{sec:sub-framework}

Figure~\ref{fig:method} illustrates the framework of Reinforced Dynamic Adversarial Training (RDAT), which aims to enhance the robustness of spatiotemporal traffic forecasting models against adversarial attacks. Our method employs dynamic selection of a subset of nodes as adversarial examples, which improves defense against dynamic attacks while reducing overfitting. To determine the optimal subset of adversarial nodes, we propose a reinforcement learning-based approach. Specifically, we formulate the node selection problem as a combinatorial optimization problem and use a policy-based network to learn the strategy that maximizes the inner loss. Our approach includes a spatiotemporal attention-based policy network that models spatiotemporal geo-distributed data, and a balanced reward function strategy to provide stable and efficient feedback to the policy network and alleviate the issue of decreasing inner loss during training. The final pre-trained policy network can be used as a node selector. To address the "forgetting issue," we also introduce a self-knowledge distillation regularization for adversarial training, where the current model is trained using knowledge distilled from the previous model's experience with adversarial attacks.

\subsection{Adversarial Training Formulation}~\label{sec:sub-ATF}
In this section, we investigate the application of traditional adversarial training methods to spatiotemporal traffic forecasting and present our proposed adversarial training formulation.

Initially, we hypothesized that protecting a larger proportion of nodes during adversarial training would lead to improved robustness of the forecasting model. However, our exploratory experiment showed that this was not the case. In fact, a higher proportion of poisoned nodes resulted in a greater degradation of the forecasting model's performance. To test this, we conducted an experiment in which we randomly selected varying proportions of nodes as adversarial samples during adversarial training, as shown in Figure~\ref{fig:Dymamic_and_static_dif_proportion}. The results of the experiment were counter-intuitive and revealed that a smaller proportion of dynamically selected nodes resulted in a more robust model.

\begin{figure}[tb]
\centering
\subfigure[80\% adversarial node]{
\begin{minipage}[]{0.5\linewidth}
\centering
\includegraphics[width=1.2in]{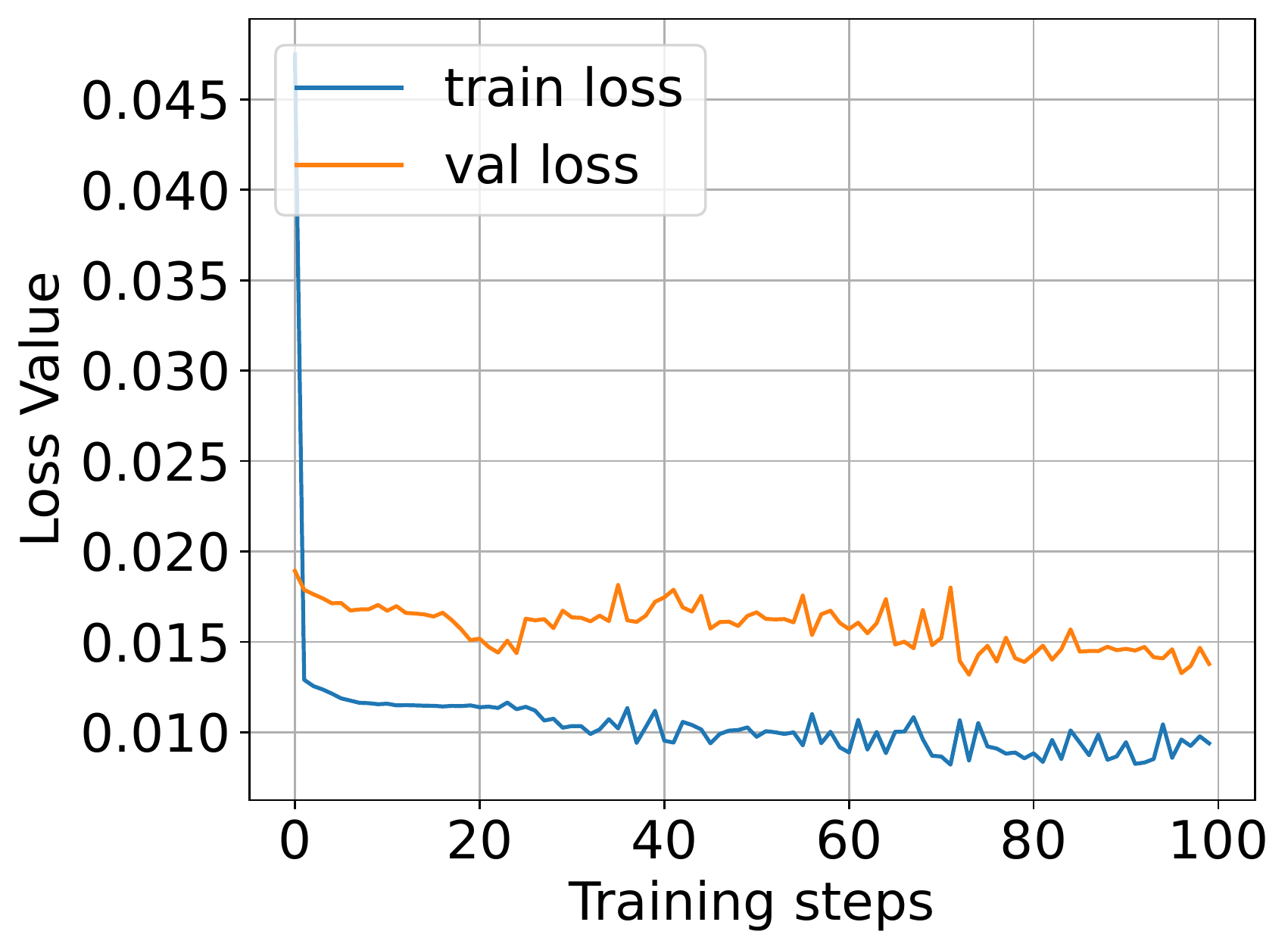}
\end{minipage}%
}%
\subfigure[10\% adversarial node]{
\begin{minipage}[]{0.5\linewidth}
\centering
\includegraphics[width=1.2in]{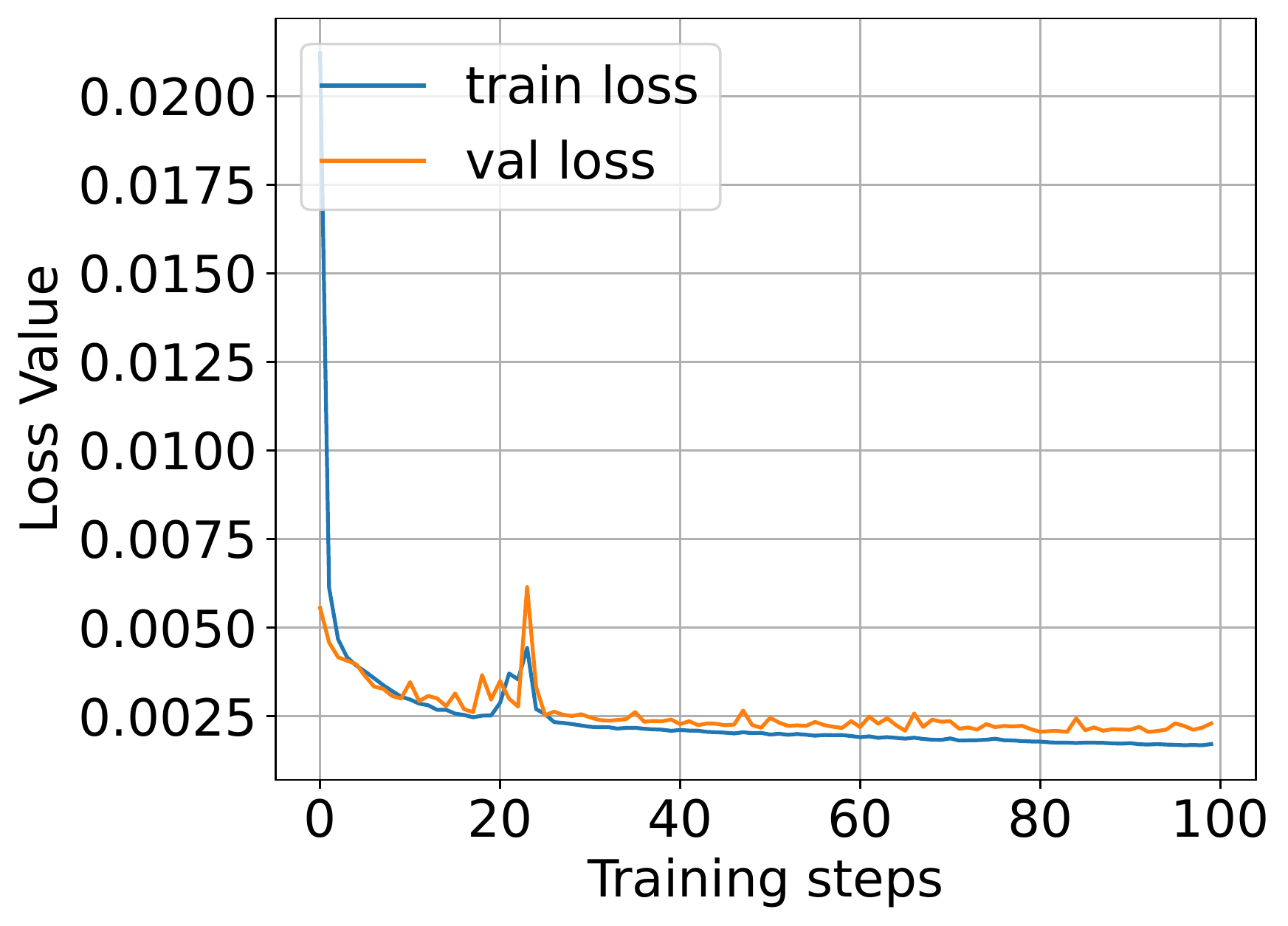}
\end{minipage}%
}%
\centering
\setlength{\belowcaptionskip}{-0.5cm}
\caption{ The figure illustrates the training loss over time. At each training step, adversarial nodes are randomly sampled as part of the training process. }\label{fig:loss_curve}
\end{figure}

Figure~\ref{fig:loss_curve} provides insight into the relationship between the proportion of adversarial nodes and the training loss. The figure shows two different scenarios, where the x-axis represents the number of training steps and the y-axis represents the training loss. In Figure~\ref{fig:loss_curve} (a), a high proportion of adversarial nodes (80\% randomly selected among all nodes) is used, resulting in overfitting and an unstable training curve. This is likely due to the model becoming too specialized to the specific set of protected nodes, leading to poor generalization to new samples. In contrast, Figure~\ref{fig:loss_curve} (b) demonstrates that a lower proportion of adversarial nodes (10\% randomly selected among all nodes) tends to mitigate the overfitting issue and results in a more stable training curve.

\textbf{Spatiotemporal adversarial examples.} Our approach is based on the insight that the key to improving the robustness of the model is to actively identify and focus on the most extreme cases of adversarial perturbation. Specifically, following~\cite{liu2021spatially}, the worst-case scenario in traffic forecasting models involves both spatial and temporal aspects. From the temporal aspect, the attacker can inject adversarial perturbations into the feature space. To effectively defend against various types of attacks, it is crucial to thoroughly explore the worst-case scenarios in the adversarial perturbation space~\cite{dong2020adversarial, madry2018towards-PGD}, which is similar to the approach taken in the field of image recognition. From a spatial perspective, it devises a dynamic node selection approach in each training epoch to maximize the inner loss and ensure that all nodes had a fair opportunity to be chosen.  To achieve this, we dynamically select a subset of nodes that exhibit spatiotemporal dependence from the full set of nodes at each training iteration. To operationalize this, we first define the permitted adversarial perturbation space as follows:
\begin{equation}
    \Psi(\mathbfcal{X'}_{t}) = \{ \mathbfcal{X}_{t} + \mathbf{\Delta}_{t} \cdot \mathbf{I}_{t} \mid \|\mathbf{I}_{t}  \| _{0} \le \eta, \| \mathbf{\Delta}_{t}|_{p}  \le \epsilon  \},
\end{equation}
where $\mathbfcal{X'}_{t}$ is the spatiotemporal adversarial example. $\mathbf{\Delta_{t}} $ is the spatiotemporal adversarial perturbations. The matrix $\mathbf{I}_{t} \in \{0,1\}^{n \times n}$ is the adversarial nodes indicator, which is a diagonal matrix whose $j$th diagonal element denotes whether or not node $v_{j}$ has been chosen as an adversarial node at time $t$. Specifically, the $j$th diagonal element of the matrix is equal to 1 if node $v_j$ has been selected as an adversarial node and 0 otherwise. The parameter $\eta$ is the budget for the number of nodes, and $\epsilon$ is the budget for the adversarial perturbation.

The adversarial training approach for spatiotemporal traffic forecasting is formulated as follows:

\begin{equation}\label{eq:ST_adversarial_training}
     \min_{\theta}\max_{\substack{\mathbfcal{X'}_{t} \in  \Psi(\mathbfcal{X'}_{t})  \\ t \in \mathcal{T}_{train}}}\sum_{t\in \mathcal{T}_{train}} \mathcal{L}_{AT}(\mathcal{F}_{\theta}(\mathbfcal{X'}_{t-\tau:t}; \mathbfcal{A} ), \mathbfcal{Y}_{t+1:t+T})   ,
\end{equation}

where $\mathbfcal{X'}_{t-\tau:t}=\{\mathbfcal{X'}_{t-\tau}, \cdots, \mathbfcal{X'}_{t} \} $ is the adversarial traffic states from time slot $t-\tau$ to $t$. $\mathcal{T}_{train} $ represents the set of time steps of all training samples. $\mathcal{L}_{AT}(\cdot)$ represents the user-specified loss function for adversarial training, which can include commonly used metrics such as Mean Squared Error (MSE) or others. The inner maximization aims to find the optimal adversarial perturbation that maximizes the loss. In the outer minimization, the model parameters are updated to minimize the prediction loss.
\subsection{Reinforced Optimal Node Subset Learning}\label{sec:sub-RONSL}
In this section, we formulate the problem of selecting the optimal subset of nodes from a set of $n$ spatiotemporal geo-distributed data sources as a combinatorial optimization problem. The problem instance, denoted as $s$, consists of $n$ nodes represented by spatiotemporal features $\mathbfcal{X}_{t-\tau:t}$ from time slot $t-\tau$ to $t$. The goal is to select $\eta$ nodes from the full set of $n$ nodes, represented by a subset of nodes $\mathbf{\Omega} = (\omega_{1}, \cdots,\omega_{\eta}) $, where $\omega_{k} \in  \{v_{1},\cdots, v_{n} \}$
and $\omega_{k} \ne \omega_{k'} \forall k \ne k'$.

Given a problem instance $s$, the objective is to learn the parameter $\phi$ of a stochastic policy $p_{\phi}( \mathbf{\Omega} \mid s)$ using the chain rule to factorize the probability of the solution. The policy network uses this information to determine the optimal subset of nodes to select in order to explore the most extreme case of adversarial perturbation at each training iteration.

\begin{equation}\label{eq:Solution}
    p_{\phi}( \mathbf{\Omega} \mid s)= \prod_{k=1}^{\eta}p_{\phi}(\omega_{k}|s,\omega _{1:k-1}).
\end{equation}
The policy network includes the encoder and decoder parts. The encoder is  to produce the geo-distributed data into the embeddings. The decoder generates the sequence of $\Omega$.

\subsubsection{Policy Network Design}

The policy network takes the spatiotemporal features $\mathbfcal{X}_{t-\tau:t}$ as input and produces the solution $\Omega$. It is composed of a spatiotemporal encoder and a multi-head-attention decoder. The encoder converts the spatiotemporal features into embeddings, and the decoder constructs the solution in an auto-regressive manner, selecting a node at a time and using the previous selection to choose the next node until the complete solution is generated.

\textbf{Spatiotemporal encoder.}
We utilized a spatiotemporal encoder, which is similar to the GraphWave Net~\cite{Graph_Wave_Net}, to transform spatiotemporal traffic data into embeddings. The spatiotemporal encoder receives spatiotemporal traffic data as input and produces embeddings of nodes as output. The spatiotemporal encoder is typically composed of multiple spatial and temporal layers.

 \textbf{Spatial layer.} We employed a self-adaptive graph convolution as the spatial layer to capture spatial dependencies. The information aggregation method was based on the diffusion model~\cite{DCRNN}, and the traffic signal was allowed to diffuse for $L$ steps. The hidden layer embedding was updated through self-adaptive graph convolution by aggregating the hidden states of neighboring nodes, 
\begin{equation}
    \mathbf{Z}_{l} = \sum_{i=0}^{L}(\mathbf{A}_{ada})^{i}\mathbf{Z'}_{l}\mathbf{W}^{i},
\end{equation}
where $\mathbf{Z'}_{l}$ denotes the outputs of hidden embedding in $l$th layers, and $\mathbf{W}^{i}$ is the model parameter for depth $i$, and  $\mathbf{A}_{ada}$ is the learnable adjacency matrix.

\textbf{Temporal layer.}
The gated temporal layer is used to process sequence data in this model. It is defined as follows,
\begin{equation}
    \mathbf{Z'}_{l} = \tanh(\vartheta_{1}\star \mathbf{E}_{l} )\odot \sigma(\vartheta_{2}\star \mathbf{E}_{l}),
\end{equation}
where $\sigma$ represents the sigmoid function, $\vartheta{1}$ and $\vartheta_{2}$ are the model parameters, $\star$ denotes the dilated convolution operation, and $\odot$ represents element-wise multiplication. $\mathbf{E}_{l}$ is the input of $l$ block and  the output of $l-1$ block. The residual links are added for each block.
\begin{equation}
    \mathbf{E}_{l+1} = \mathbf{Z}_{l} + \mathbf{E}_{l}
\end{equation}

The hidden states  of different  layers are concatenated   and passed into two  multilayer perceptions (MLP) to get the final node embeddings. 
\begin{equation}
    \mathbf{F} = MLP(\|_{l=1}\mathbf{Z}_{l}), 
\end{equation}
where $\mathbf{F}$ is the set of node embeddings, and $\mathbf{F}_{i}$ is the embedding for node $v_{i}$. The average of all the node embeddings is denoted as the graph embedding and can be represented as $\frac{1}{n}\sum_{i} \mathbf{F}_{i}$.

\textbf{Multi-Head-Attention decoder.}
The decoder  generates a sequence of nodes, $\Omega$, by iteratively selecting individual nodes, $\omega_k$, at each step $k$, using both the encoder's embedding and the output of previous steps, $\omega_{k'}$, for $k' < k$, as input.

Specifically, the decoder's input comprises the graph embedding and the embedding of the last node, with the first selected node's embedding being a learned embedding. The decoder calculates the probability of each node being selected as an adversarial node, while also taking computational efficiency into consideration. During decoding process, context is represented by a special context node (c). To this end, we incorporate the attention-based decoder~\cite{kool2018attention} to compute an attention layer on top of the decoder, with messages only to the context node (c). The context node   embedding is defined as follows:

\begin{equation}
\mathbf{U}_{(c)}=
\left\{\begin{matrix}
[\mathbf{\bar{U}},\mathbf{v} ]  &  k =1 \\
[\mathbf{\bar{U}},\mathbf{U}_{\omega_{k-1}}] & k >1,
\end{matrix}\right.
\end{equation}

where $\mathbf{\bar{U}} = \frac{1}{n}\sum_{i} \mathbf{F}_{i}$ is graph embedding and $\mathbf{v}$ is a learned embedding at the first iterative step. $\mathbf{U}{\omega_{k-1}}$ is the embedding of the last selected node at the k-1-th iterative step.

To update the context node embedding for message information, the new context node embedding is calculated by Multi-head-attention:

\begin{equation}
    U'_{(c)}= \|_{j=1} MHA_{j}( \mathbf{q}_{(c)},\mathbf{k}_{j},\mathbf{v}_{j})
\end{equation}

Where $MHA_{j}(\mathbf{q}_{(c)},\mathbf{k}_{j},\mathbf{v}_{j})= softmax(\frac{\mathbf{q}_{(c)}\mathbf{k}_{j}}{\sqrt{d_{k}}})\mathbf{v}_{j}$ is self attention,  and  $\mathbf{q}_{(c)} = \mathbf{W}^{Q}_{(c)}\mathbf{U}_{(c)}$, $\mathbf{k}_{j}=W^{K}_{j}\mathbf{H'} $, and  $\mathbf{v}_{j}=W^{V}_{j}\mathbf{H'}$ $(j=1,\cdots,M)$.

To compute the next node probability,  the keys and values come from the initial node embedding 
\begin{equation}
  \mathbf{q}= \mathbf{W}^{Q}\mathbf{U'}_{(c)}, \quad   \mathbf{k}_{i}=W^{K}\mathbf{H'}_{i}.
\end{equation}

We first compute the logit with a single attention head, use the new context node to query with all nodes,
\begin{equation}
    r_{j}=\left\{\begin{matrix}
C\tanh(\frac{\mathbf{q}^{T}\mathbf{k}_{j} }{\sqrt{d_{k}} } )  & \text{if} \quad j\ne \omega_{k'} \quad\forall  k'<k  \\
  -\infty & \quad \text{otherwise} 
\end{matrix}\right. 
\end{equation}
where $C$ is the constant, and  the selected node is masked with $r_{j} = -\infty$.

Then, the final probability of nodes using the chain rule and the softmax function, the probability of each node can be calculated based on the softmax function,
\begin{equation}
p_{i}=p_{\theta }(\omega_{k}=i \mid s,\omega_{1:k-1})=\frac{e^{r_{i}}}{\sum_{j}e^{r_{j}}} 
\end{equation}
where $p_{i}$ is the probability of node $i$ and $\omega_{k}$ is the current node. The node with the highest probability among all nodes is selected as the next sampling node.

\subsubsection{Balanced-reward Function Design } 
The main challenge in policy network learning is evaluating the solutions $\Omega$ generated by the policy network. One approach is to use the inner loss, computed using the solutions $\Omega$, as the reward, with larger values indicating better solutions. However, as the training progresses, the inner loss is expected to decrease as the model becomes more robust, which can lead to incorrect feedback and suboptimal solutions. To address this, we propose a balanced strategy for the reward function. Instead of solely using the inner loss, we compare the results generated by the policy network to those generated by a baseline node selector and use the difference as the reward. This approach provides stable and efficient feedback to the policy network and helps to mitigate the issue of decreasing inner loss during training.

Specifically, we first obtain the set of adversarial nodes indicator $\mathbf{I}_{t}$ based on the solution  $\mathbf{\Omega} = (\omega_{1}, \cdots,\omega_{K}) $, where $\omega_{k} \in \{v_{1},\cdots, v_{n} \}$, using the following function:
\begin{equation}\label{eq:target_node_mask}
\mathbf{I}_{(i,i),t}=\begin{cases}
 1 & \text{ if } v_i \in \mathbf{\Omega} \\
 0 & \text{ otherwise }, \end{cases}   
\end{equation}
where $I_{(i,i),t}$ denotes the $i$-th diagonal element of $\mathbf{I}_t$ at time step $t$.

Instead of using gradient-based methods to calculate the adversarial examples, we directly sample a random variable $\mathbf{\Delta}$ from a probability distribution $\pi(\mathbf{\Delta})$ to calculate the adversarial examples in Eq~\ref{eq:ST_adversarial_training} for computational efficiency.

\begin{equation}
   \mathcal{X'}_{t} = \mathcal{X}_{t}+\mathbf{\Delta} \cdot \mathbf{I}_{t}
\end{equation}
In the implementation, we choose the uniform distribution with range $(-\epsilon, \epsilon)$ as the source of perturbations $\mathbf{\Delta}$.

To evaluate the performance of our forecasting model when using the nodes in the solution as adversarial nodes, we calculate the cost function as follows:

\begin{equation}
   L(\Omega) =\mathcal{L}_{MSE}(\mathcal{F}_{\theta}(\mathcal{X'}_{t-\tau:t}; \mathcal{A} ),\mathbfcal{Y}_{t+1:t+T} ) 
\end{equation}
where $\mathcal{L}_{MSE}(\cdot)$ is the MSE loss, and $\mathcal{X'}_{t-\tau}=\{\mathcal{X'}_{t-\tau},\cdots,\mathcal{X'}_{t}\}$.

To ensure that the policy network receives stable and efficient feedback, we implement a balanced strategy for the reward. Specifically, we use a baseline node selector $B$ (\eg Random selector, randomly select the nodes, \etc ) to select nodes as the solution $\Omega_{b}$. The results generated by the policy network are then compared to the baseline results, and the difference is used as the reward. This is represented by the following equation:
\begin{equation}\label{eq:balanced_reward}
 r(\Omega^{(p)}) = L(\Omega^{(p)}) -    L(\Omega^{(b)}) 
\end{equation}
Where $\Omega^{(p)}$ is the solution generated by policy network, and $\Omega^{(b)}$ is the solution generated by the baseline selector,    we use the superscript (p) and (b) to align with the  policy network selector and baseline selector, respectively.  In this way, the balanced-reward function $r(\Omega^{(p)})$ is used as the reward signal to guide the policy network to update the solution $\mathbf{\Omega}$. In practice, we adopt a  heuristic approach  as a baseline selector  to select nodes named TNDS~\cite{liu2021spatially}.

\begin{algorithm}[t]
\caption{Reinforced Optimal Node Subset Learning}\label{alg:policy_network_learning}
\KwIn{
 Traffic data, number of epochs E, maximum perturbation budget $\epsilon$, adversarial node budget $\eta$, baseline
selector B, and inner samples iterations $b$.
}
\KwResult{ Policy network.}
Initialize model paramaters $\theta $ and $\phi$\;
{\For { $i=1$  to $E$ }
{
{\For { $j=1$  to $b$ }
{Obtain the solution  $\Omega^{(p)}$ by Equation~\ref{eq:Solution} \;  
Generate the baseline solution   $\Omega^{(b)}$ by baseline selector B \;  
Compute the balanced reward by Equation~\ref{eq:balanced_reward} \;
Compute the policy network loss by Equation~\ref{eq:policy_network_loss} \;
 Update $\phi$ using Adam optimization with the gradients of $\mathcal{L}(\phi \mid s)$ \;
}}
Generate perturbed adversarial features $\mathbf{{X}'}_{t-\tau:t}$ by Equation~\ref{eq:white-box_ST-PGD}\;  
Compute the forecasting loss by Equation~\ref{eq:forecasting_loss}\;
 Update $\theta$ using Adam optimization with the gradients of $\mathcal{L}(\theta)$ \;
}}
Return $p_{\phi }(\Omega \mid s)$.
\end{algorithm}

\subsubsection{Policy Network Training}

The training of the policy network is executed by alternately training the policy network and the spatiotemporal traffic forecasting model in an adversarial manner. Specifically, the policy network generates a solution sequence, denoted as $\Omega$, based on the input $\mathbfcal{X}_{t-\tau:t}$. The balanced-reward is then calculated and used to update the policy network. Subsequently, the final node selection indicators are employed to calculate the adversarial examples, denoted as $\mathbfcal{X'}_{t-\tau:t}$, via Projected Gradient Decent (PGD)~\cite{madry2018towards-PGD} according to the following:
\begin{equation}\label{eq:white-box_ST-PGD}
\begin{aligned}
\mathbfcal{X'}_{t-\tau:t}^{(i)} =&\text{clip}_{\mathbfcal{X'}_{t-\tau:t}, \epsilon }(\mathbfcal{X'}_{t-\tau:t}^{(i-1)} \\ 
& +\gamma  \text{sign}( \nabla \mathcal{L}_{MSE}(\mathcal{F}_{\theta}(\mathcal{X'}_{t-\tau:t}; \mathcal{A} ),\mathbfcal{Y}_{t+1:t+T} ) \cdot \mathbf{I}_{t})),
\end{aligned}
\end{equation}
where $\text{clip}_{\mathbfcal{X'}_{t-\tau:t}, \epsilon }(\cdot)$ operator is used to limit the maximum perturbation of the variable $\mathbfcal{X'}_{t-\tau:t}$ to a budget of $\epsilon$. The adversarial examples at the $i$th iteration is represented by $\mathbfcal{X'}_{t-\tau:t}^{(i)}$.
$\gamma $ is the step size, and $\mathbf{I}_{t}$ is the final node selection indicators obtained from the policy network, and $\mathcal{L}_{MSE}(\cdot)$ is the mean squared error loss function.

Subsequently, the spatiotemporal traffic forecasting model is trained on the adversarial examples to optimize the forecasting model loss as follows, \begin{equation}\label{eq:forecasting_loss}
  \mathcal{L}_{F_{\theta}} = \mathcal{L}_{MSE}(\mathcal{F}_{\theta}(\mathcal{X'}_{t-\tau:t}; \mathcal{A} ),\mathbfcal{Y}_{t+1:t+T} )\end{equation}

To train the policy network, the loss function is defined as shown in Equation \ref{eq:policy_network_loss}.
\begin{equation}\label{eq:policy_network_loss}
   \mathcal{L}(\phi \mid s ) = \mathbb{E}_{ p_{\phi}( \mathbf{\Omega}^{(p)} \mid s)}[(r(\Omega^{(p)}) + c)  \nabla \log p_{\phi}( \mathbf{\Omega}^{(p)} \mid s)] , 
\end{equation}
Where $c$ is a constant. The policy network is optimized using gradient descent with the REINFORCE algorithm~\cite{Williams92}, and the Adam optimizer is used, as outlined in Algorithm~\ref{alg:policy_network_learning}.

\subsection{ Adversarial Training with Regularization}\label{sec:sub-ATR}
Another challenge in adversarial training for spatiotemporal traffic forecasting is instability, which can occur when the adversarial nodes are constantly changing during the training process. This can lead to a situation where the model is unable to effectively remember all the historical adversarial nodes, resulting in a lack of robustness against stronger attack strengths, commonly referred to as the "forgetting issue"~\cite{DongXYPDSZ22, Yu0SY0GL22}. To address this, we propose using knowledge distillation (KD) to transfer knowledge from a teacher model to a student model. Previous studies have shown that KD can improve the adversarial robustness of models~\cite{GoldblumFFG20, abs-2203-07159}. However, traditional teacher models are static and cannot provide dynamic knowledge. To overcome this limitation, we introduce a new self-knowledge distillation regularization for adversarial training. Specifically, we use the model from the previous epoch as the teacher model, meaning that the current spatiotemporal traffic forecasting model is trained using knowledge distilled from the previous model. In this way, the current model can learn from the previous model's experience with adversarial attacks. The knowledge distil loss is defined as follows,

\begin{equation}
    \mathcal{L}_{KD} =   \mathcal{L}_{MSE}(\mathcal{F}_{\theta}(\mathbfcal{X'}_{t-\tau:t} ; \mathbfcal{A}),\mathcal{F}^{T}_{\theta}(\mathbfcal{X}_{t-\tau:t}; \mathbfcal{A}))
\end{equation}
where $\mathcal{L}_{KD}(\cdot)$ is the knowledge distillation loss (\eg. MSE \etc). and $\mathcal{F}^{T}_{\theta}(\cdot)$ is the teacher model, which is adopted from last trained model. In summary, the final adversarial training loss is defined as follows,

\begin{equation}
\begin{aligned}\label{eq:AT_loss}
\mathcal{L}_{AT}  = &     \mathcal{L}_{MSE}(\mathcal{F}_{\theta}(\mathbfcal{X'}_{t-\tau:t}; \mathbfcal{A} ), \mathbfcal{Y}_{t+1:t+T})\\ 
&  + \alpha \mathcal{L}_{KD}( \mathcal{F}_{\theta}(\mathbfcal{X'}_{t-\tau:t} ; \mathbfcal{A}),\mathcal{F}^{T}_{\theta}(\mathbfcal{X}_{t-\tau:t}; \mathbfcal{A}))
\end{aligned}
\end{equation}
where $\alpha$ is a parameter that controls the amount of knowledge transferred from the teacher model. Note that in the first training epoch, the $\mathcal{L}_{MSE}(\cdot)$ function is used directly as the adversarial training loss.

\subsubsection{Adversarial Training on Spatiotemporal Traffic Forecasting Model }
The training process is divided into two stages. In the first stage, we train the policy network using Algorithm~\ref{alg:policy_network_learning}. In the second stage, we use the pre-trained policy network to select the adversarial nodes for computational efficiency, and then compute the adversarial examples by using the PGD method. The adversarial examples are computed using Equation~\ref{eq:white-box_ST-PGD} with the adversarial training loss $\mathcal{L}_{AT}$ in Equation~\ref{eq:AT_loss}. Finally, we update the forecasting model parameters $\theta$ using the Adam optimizer. The entire training process is outlined in Algorithm~\ref{alg:adversarial_learning} in Appendix~\ref{sec:appdix-alg}.
\eat{

\begin{equation}\label{eq:ST-PGD}
\begin{aligned}
\mathbfcal{X'}_{t-\tau:t}^{(i)} =&\text{clip}_{\mathbfcal{X'}_{t-\tau:t}, \epsilon }(\mathbfcal{X'}_{t-\tau:t}^{(i-1)} \\ 
& +\alpha \text{sign}( \nabla \mathcal{L}_{AT}(\mathcal{F}_{\theta}(\mathcal{X'}_{t-\tau:t}; \mathcal{A} ),\mathbfcal{Y}_{t+1:t+T} ) \cdot I_{t})).
\end{aligned}
\end{equation}
In the end, we use the Adam to update the  forecasting model parameter $\theta$. The overall all training  process is outlined in Algorithm~\ref{alg:adversarial_learning}. 
}

\section{Experiments}
In this section, we compare the performance of our proposed method to state-of-the-art adversarial training methods on spatiotemporal traffic forecasting. We aim to answer the following evaluation questions :
\begin{itemize}
    \item \textbf{EQ1} Does the proposed method improve the adversarial robustness of spatiotemporal traffic forecasting models on real-world datasets?
    \item \textbf{EQ2} How effective are the adversarial training framework, policy network, and self-distillation module?
    \item \textbf{EQ3} How robust is the proposed method with respect to different hyperparameter values?
\end{itemize}

To answer these questions, we present the datasets, baselines, target model, evaluation metric, and implementation details, followed by experiments.
\subsection{Evaluation}
\textbf{Datasets.}
We use two real-world spatiotemporal traffic forecasting datasets, PEMS-BAY~\cite{DCRNN} and PEMS-D4~\cite{DCRNN}, which were collected by the California Performance of Transportation (PeMS) and contain traffic speed and flow data, respectively. These datasets are sorted by time in ascending order and have a time interval of 5 minutes between consecutive points. We allocate 70\% of the data for training, 10\% for validation, and 20\% for testing.

\textbf{Baselines.}
There are few studies in the current literature that can be directly applied to the real-value traffic forecasting defense setting. Therefore, to ensure fair comparisons, we use state-of-the-art adversarial training methods, including adversarial training (AT)~\cite{madry2018towards-PGD}, TRADE~\cite{tramer2019adversarial-TRADE}, Mixup~\cite{zhang2018mixup}, and GrpahAT~\cite{FengHTC21} with the random node selection strategy. Additionally, we use the recent state-of-the-art traffic forecasting attack method, TDNS~\cite{liu2022practical}, as a dynamic node selection method in combination with AT, which we refer to as AT-TDNS. These methods serve as our baselines.

\textbf{Target model.} We adopt the state-of-the-art spatiotemporal traffic forecasting model, GraphWave Net~\cite{Graph_Wave_Net}, as the target model to evaluate the generalization of our adversarial training framework. Results on additional target models are included in the appendix~\ref{sec:ARC}.

\textbf{Evaluation metrics.}
For evaluating the adversarial robustness of spatiotemporal traffic forecasting models, we adopt the mean absolute error (MAE), and root mean squared error (RMSE) as evaluation metrics.

\textbf{Implement details.}
We conduct our experiments using Pytorch on a Linux Centos Server with 12 RTX 3090 GPUs and 2 RTX A40 GPUs. The traffic data is normalized to the range [0,1], and the input and output lengths are set to $\tau=12$ and $T=12$, respectively.  We follow the attack setting in~\cite{liu2022practical}, using PGD-Random, PGD-PR, PGD-Centrality, PGD-Degree, and PGD-TNDS as the attacker. The regularization parameter $\alpha$ is set to $0.4$. The perturbation magnitude $\epsilon$ is 0.5 for both training and testing. During training, we select 10\% of the total nodes as adversarial examples at each epoch, while in testing, we use a stronger attack strength and select 20\% of the total nodes as adversarial examples. We conducted the experiments five times and present the average results along with the standard deviations (STD) of the metrics.

\begin{figure}[t]
\centering
\subfigure[PeMS-BAY]{
\begin{minipage}[]{0.5\linewidth}
\centering
\includegraphics[width=1.2in]{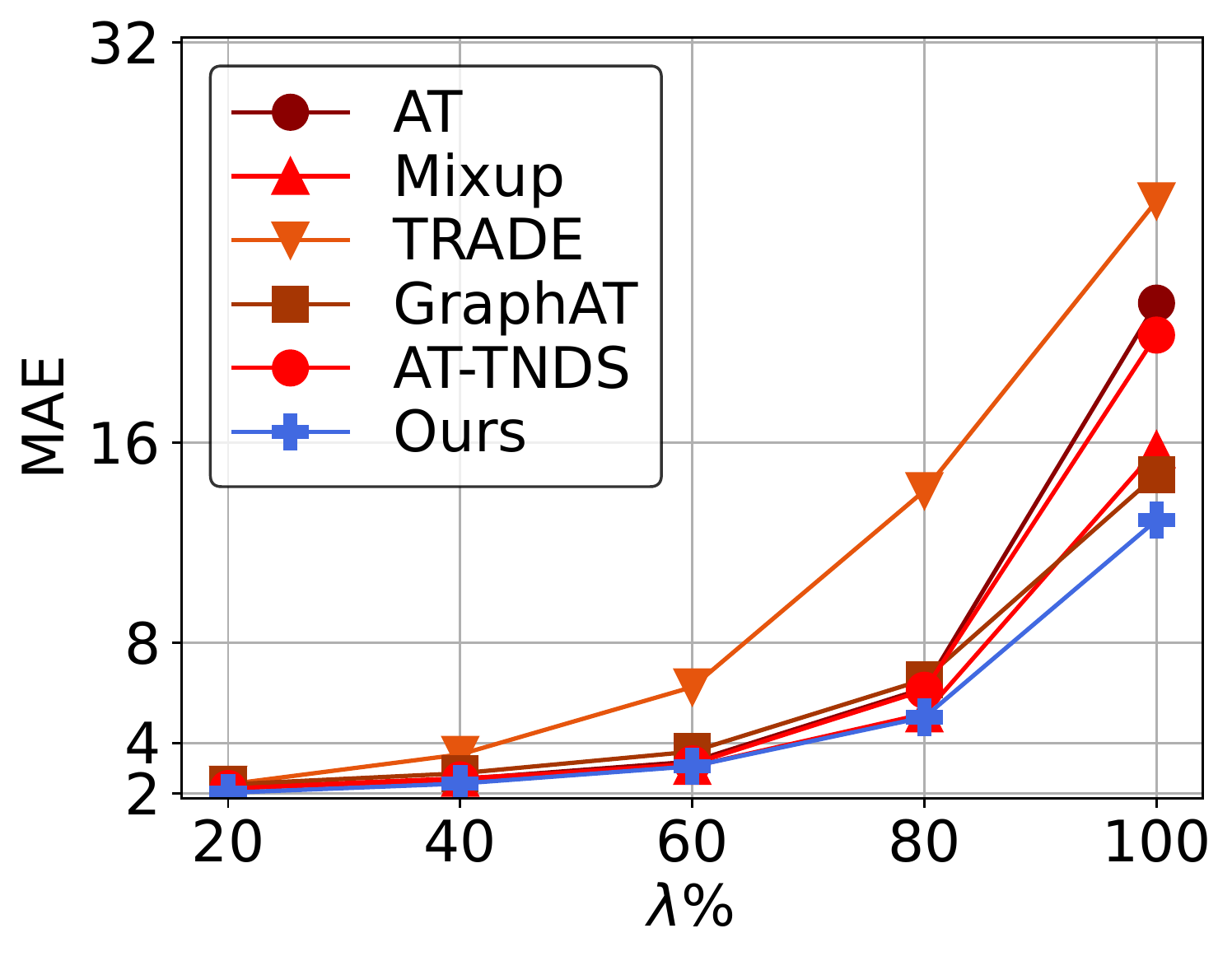}
\end{minipage}%
}%
\subfigure[PeMS-D4]{
\begin{minipage}[]{0.5\linewidth}
\centering
\includegraphics[width=1.2in]{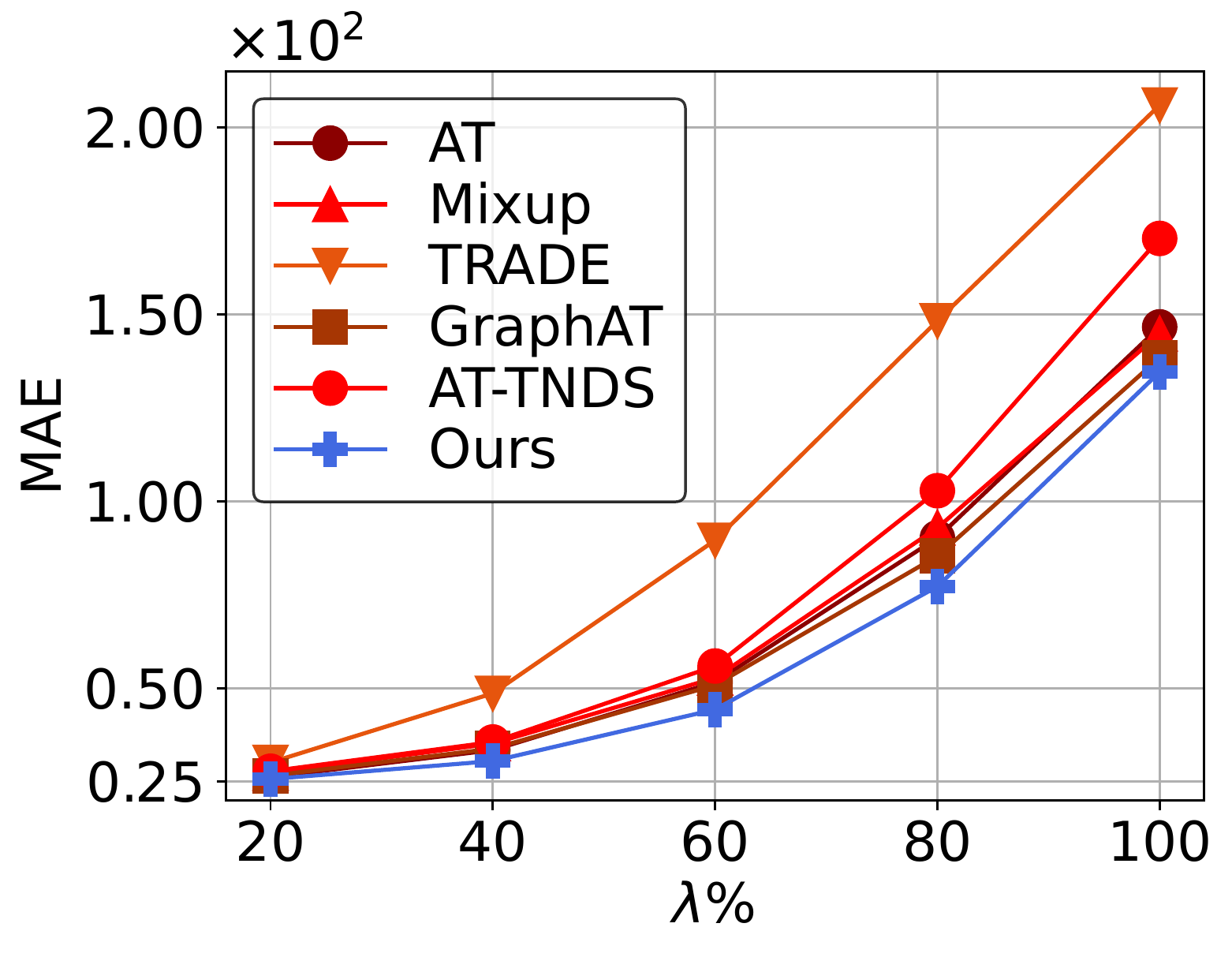}
\end{minipage}%
}%
\centering
\setlength{\belowcaptionskip}{-0.7cm}
\caption{ Adversarial robustness performance under different attack strengths.}\label{fig:PGD_Random}
\end{figure}

\subsection{EQ1:  Main Results}

To answer the first evaluation question (EQ1), we compare the proposed method with state-of-the-art adversarial training methods on two real-world traffic forecasting datasets, PEMS-BAY and PEMS-D4. Table \ref{tab:main_results_PeMS-BAY} and \ref{tab:main_results_PeMS-D4} present the overall adversarial robustness performance of our proposed defense method against adversarial traffic forecasting attacks and five baselines, evaluated using two metrics. Our adversarial training method significantly improves the adversarial robustness of traffic forecasting models, as shown by the (68.55\%, 66.0350\%) and (69.70\%, 69.0343\%) improvements in the PeMS-BAY and PEMS-D4 datasets, respectively, under the PGD-Random attack where no defense strategy was applied. Additionally,  our method achieved a (1.12\%, 2.10\%) improvement in clean performance and (7.65\%, 12.19\%), (7.75\%, 3.31\%), and (7.35\%, 2.81\%) improvements in adversarial robustness under the PGD-Random, PGD-PR, PGD-Degree attacks, respectively, compared to almost all baselines in terms of PEMS-BAY. While our method was slightly weaker than GraphAT (4.4545 compared to 4.3762) in terms of RMSE under the PGD-TNDS attack, our method had lower standard deviation, indicating increased stability. Overall, our method significantly enhances the adversarial robustness of the traffic forecasting model against adversarial traffic attacks.

\begin{table*}[t]
\caption{Adversarial robustness performance ( MAE (std) / RMSE (std)) on dataset PEMS-BAY}\label{tab:main_results_PeMS-BAY}
\scalebox{0.85}{
\begin{tabular}{c|clclclclclcl}
\hline
Method & \multicolumn{2}{c}{Non-attack}                                                                          & \multicolumn{2}{c}{PGD-Random}                                                                          & \multicolumn{2}{c}{PGD-PR}                                                                              & \multicolumn{2}{c}{PGD-Centrality}                                                                      & \multicolumn{2}{c}{PGD-Degree}                                                                          & \multicolumn{2}{c}{PGD-TNDS}                                                                            \\ \hline
Non-defense                 & \multicolumn{2}{c}{\begin{tabular}[c]{@{}c@{}}\textbf{1.7811(0.0528 )}/\\ \textbf{3.7643 (0.0870)}\end{tabular}}          & \multicolumn{2}{c}{\begin{tabular}[c]{@{}c@{}}6.4196 (0.8209)/\\ 12.0430 (1.5761)\end{tabular}}         & \multicolumn{2}{c}{\begin{tabular}[c]{@{}c@{}}6.4600 (0.9506)/\\ 12.2761 (1.8196)\end{tabular}}         & \multicolumn{2}{c}{\begin{tabular}[c]{@{}c@{}}6.4488 (0.8349)/\\ 12.1571 (1.6495)\end{tabular}}         & \multicolumn{2}{c}{\begin{tabular}[c]{@{}c@{}}6.4426 (0.8360)/\\ 12.141 (1.6045)\end{tabular}}          & \multicolumn{2}{c}{\begin{tabular}[c]{@{}c@{}}7.8215 (0.7425)/\\ 13.4806 (1.0977)\end{tabular}}         \\ \hline
AT                          & \multicolumn{2}{c}{\begin{tabular}[c]{@{}c@{}}1.8867 (0.1507)/\\ 3.8725 (0.2927)\end{tabular}}          & \multicolumn{2}{c}{\begin{tabular}[c]{@{}c@{}}2.1944 (0.1528)/\\ 4.2984 (0.2707)\end{tabular}}          & \multicolumn{2}{c}{\begin{tabular}[c]{@{}c@{}}2.2261 (0.1447)/\\ 4.3447 (0.2758)\end{tabular}}          & \multicolumn{2}{c}{\begin{tabular}[c]{@{}c@{}}2.1790 (0.1352)/\\ 4.2747 (0.2484)\end{tabular}}          & \multicolumn{2}{c}{\begin{tabular}[c]{@{}c@{}}2.2155 (0.1589)/\\ 4.3229 (0.2918)\end{tabular}}          & \multicolumn{2}{c}{\begin{tabular}[c]{@{}c@{}}2.1968 (0.1485)/\\ 4.4661 (0.2887)\end{tabular}}          \\ \hline
Mixup                       & \multicolumn{2}{c}{\begin{tabular}[c]{@{}c@{}}1.9828 (0.1301)/\\ 3.8759 (0.1283)\end{tabular}}          & \multicolumn{2}{c}{\begin{tabular}[c]{@{}c@{}}2.2168 (0.1277)/\\ 4.2505 (0.1151)\end{tabular}}          & \multicolumn{2}{c}{\begin{tabular}[c]{@{}c@{}}2.2418 (0.1278)/\\ 4.3034 (0.1220)\end{tabular}}          & \multicolumn{2}{c}{\begin{tabular}[c]{@{}c@{}}2.2198 (0.1307)/\\ 4.2677 (0.1186)\end{tabular}}          & \multicolumn{2}{c}{\begin{tabular}[c]{@{}c@{}}2.2198 (0.1307)/\\ 4.2677 (0.1186)\end{tabular}}          & \multicolumn{2}{c}{\begin{tabular}[c]{@{}c@{}}2.2961(0.1257)/\\ 4.5534 (0.3081)\end{tabular}}           \\ \hline
TRADE                       & \multicolumn{2}{c}{\begin{tabular}[c]{@{}c@{}}1.8196 (0.1748)/\\ 3.8587 (0.2813)\end{tabular}}          & \multicolumn{2}{c}{\begin{tabular}[c]{@{}c@{}}2.3555 (0.3175)/\\ 4.5893 (0.4351)\end{tabular}}          & \multicolumn{2}{c}{\begin{tabular}[c]{@{}c@{}}2.3923 (0.3355)/\\ 4.6444 (0.4315)\end{tabular}}          & \multicolumn{2}{c}{\begin{tabular}[c]{@{}c@{}}2.4044 (0.3270)/\\ 4.6587 (0.4239)\end{tabular}}          & \multicolumn{2}{c}{\begin{tabular}[c]{@{}c@{}}2.3787 (0.6469)/\\ 4.6103 (0.4143)\end{tabular}}          & \multicolumn{2}{c}{\begin{tabular}[c]{@{}c@{}}2.9446 (0.6469)/\\ 5.7703 (0.9296)\end{tabular}}          \\ \hline
GraphAT                     & \multicolumn{2}{c}{\begin{tabular}[c]{@{}c@{}}2.0170 (0.3122)/\\ 3.8210 (0.0655)\end{tabular}}          & \multicolumn{2}{c}{\begin{tabular}[c]{@{}c@{}}2.3512 (0.3855)/\\ 4.2395 (0.1763)\end{tabular}}          & \multicolumn{2}{c}{\begin{tabular}[c]{@{}c@{}}2.3932 (0.3784)/\\ 4.2943 (0.1548)\end{tabular}}          & \multicolumn{2}{c}{\begin{tabular}[c]{@{}c@{}}2.3526 (0.3847)/\\ 4.2338 (0.1753)\end{tabular}}          & \multicolumn{2}{c}{\begin{tabular}[c]{@{}c@{}}2.3721 (0.3705)/\\ 4.2576 (0.1522)\end{tabular}}          & \multicolumn{2}{c}{\begin{tabular}[c]{@{}c@{}}2.3223 (0.3727)/\\ \textbf{4.3762 (0.1933)} \end{tabular}}          \\ \hline
AT-TNDS                     & \multicolumn{2}{c}{\begin{tabular}[c]{@{}c@{}}1.9405 (0.1274)/\\ 3.8964 (0.1433)\end{tabular}}          & \multicolumn{2}{c}{\begin{tabular}[c]{@{}c@{}}2.1864 (0.1248)/\\ 4.2190 (0.0782)\end{tabular}}          & \multicolumn{2}{c}{\begin{tabular}[c]{@{}c@{}}2.2343 (0.1280)/\\ 4.3064 (0.0818)\end{tabular}}          & \multicolumn{2}{c}{\begin{tabular}[c]{@{}c@{}}2.1871 (0.1185)/\\ 4.2282 (0.0807)\end{tabular}}          & \multicolumn{2}{c}{\begin{tabular}[c]{@{}c@{}}2.2055 (0.1197)/\\ 4.2416 (0.0800)\end{tabular}}          & \multicolumn{2}{c}{\begin{tabular}[c]{@{}c@{}}2.2263 (0.1466)/\\ 4.4924 (0.2122)\end{tabular}}          \\ \hline
Ours                        & \multicolumn{2}{c}{\begin{tabular}[c]{@{}c@{}}1.7992 (0.0781)/\\ 3.7777 (0.1277)\end{tabular}} & \multicolumn{2}{c}{\textbf{\begin{tabular}[c]{@{}c@{}}2.0191 (0.0556)/\\ 4.0904 (0.1212)\end{tabular}}} & \multicolumn{2}{c}{\textbf{\begin{tabular}[c]{@{}c@{}}2.0611 (0.0503)/\\ 4.1609 (0.1019)\end{tabular}}} & \multicolumn{2}{c}{\textbf{\begin{tabular}[c]{@{}c@{}}2.0245 (0.0546)/\\ 4.1009 (0.1101)\end{tabular}}} & \multicolumn{2}{c}{\textbf{\begin{tabular}[c]{@{}c@{}}2.0433 (0.0521)/\\ 4.1222 (0.1000)\end{tabular}}} & \multicolumn{2}{c}{\begin{tabular}[c]{@{}c@{}}\textbf{2.0944 (0.0738)}/\\ 4.4545 (0.1628)\end{tabular}} \\ \hline
\end{tabular}}
\end{table*}

\begin{table*}[t]
\caption{Adversarial robustness performance ( MAE (std) / RMSE (std)) on dataset PEMS-D4}\label{tab:main_results_PeMS-D4}
\scalebox{0.85}{
\begin{tabular}{c|clclclclclcl}
\hline
Method & \multicolumn{2}{c}{Non-attack}                                                                            & \multicolumn{2}{c}{PGD-Random}                                                                            & \multicolumn{2}{c}{PGD-PR}                                                                                & \multicolumn{2}{c}{PGD-Centrality}                                                                        & \multicolumn{2}{c}{PGD-Degree}                                                                            & \multicolumn{2}{c}{PGD-TNDS}                                                                              \\ \hline
Non-defense                 & \multicolumn{2}{c}{\begin{tabular}[c]{@{}c@{}}22.4184 (0.4692)/\\ 34.6917 (0.6899)\end{tabular}}          & \multicolumn{2}{c}{\begin{tabular}[c]{@{}c@{}}84.7836 (8.1358)/\\ 126.4276 (10.3251)\end{tabular}}        & \multicolumn{2}{c}{\begin{tabular}[c]{@{}c@{}}85.2586 (8.8071)/\\ 128.5284 (10.5656)\end{tabular}}        & \multicolumn{2}{c}{\begin{tabular}[c]{@{}c@{}}83.3945 (8.8071)/\\ 125.7175 (11.4477)\end{tabular}}        & \multicolumn{2}{c}{\begin{tabular}[c]{@{}c@{}}85.2536 (8.8048)/\\ 129.5471 (11.1326)\end{tabular}}        & \multicolumn{2}{c}{\begin{tabular}[c]{@{}c@{}}106.1603 (13.5806)/\\ 157.1661 (16.2174)\end{tabular}}      \\ \hline
AT                          & \multicolumn{2}{c}{\begin{tabular}[c]{@{}c@{}}23.6584 (1.0385)/\\ 36.1477 (1.2730)\end{tabular}} & \multicolumn{2}{c}{\begin{tabular}[c]{@{}c@{}}26.1930 (1.2662)/\\ 40.0484 (1.7344)\end{tabular}}          & \multicolumn{2}{c}{\begin{tabular}[c]{@{}c@{}}26.4923 (1.0598)/\\ 40.4635 (1.5168)\end{tabular}}          & \multicolumn{2}{c}{\begin{tabular}[c]{@{}c@{}}26.1568 (1.0598)/\\ 40.4635 (1.5168)\end{tabular}}          & \multicolumn{2}{c}{\begin{tabular}[c]{@{}c@{}}26.6188 (1.1972)/\\ 41.4411 (1.5931)\end{tabular}}          & \multicolumn{2}{c}{\begin{tabular}[c]{@{}c@{}}28.1043 (1.2327)/\\ 43.7800 (1.0946)\end{tabular}}          \\ \hline
Mixup                       & \multicolumn{2}{c}{\begin{tabular}[c]{@{}c@{}}24.4370 (0.6236)/\\ 36.7487 (0.2843)\end{tabular}}          & \multicolumn{2}{c}{\begin{tabular}[c]{@{}c@{}}27.3264 (0.5640)/\\ 41.9264 (0.6865)\end{tabular}}          & \multicolumn{2}{c}{\begin{tabular}[c]{@{}c@{}}27.6455 (0.5097)/\\ 42.8240 (0.7187)\end{tabular}}          & \multicolumn{2}{c}{\begin{tabular}[c]{@{}c@{}}28.8220 (3.5680)/\\ 44.4667 (5.2713)\end{tabular}}          & \multicolumn{2}{c}{\begin{tabular}[c]{@{}c@{}}27.9851 (0.4878)/\\ 43.6414 (0.7261)\end{tabular}}          & \multicolumn{2}{c}{\begin{tabular}[c]{@{}c@{}}29.0781 (0.9618)/\\ 45.1856 (0.8442)\end{tabular}}          \\ \hline
TRADE                       & \multicolumn{2}{c}{\textbf{\begin{tabular}[c]{@{}c@{}}22.1531 (1.1497)/\\ 34.1600 (1.4682)\end{tabular}}} & \multicolumn{2}{c}{\begin{tabular}[c]{@{}c@{}}30.1296 (1.5853)/\\ 46.6367 (2.5212)\end{tabular}}          & \multicolumn{2}{c}{\begin{tabular}[c]{@{}c@{}}30.2241 (2.0172)/\\ 47.2981 (3.0243)\end{tabular}}          & \multicolumn{2}{c}{\begin{tabular}[c]{@{}c@{}}30.4041 (1.6338)/\\ 47.8004 (2.8228)\end{tabular}}          & \multicolumn{2}{c}{\begin{tabular}[c]{@{}c@{}}30.7245 (2.0601)/\\ 48.5267 (3.1621)\end{tabular}}          & \multicolumn{2}{c}{\begin{tabular}[c]{@{}c@{}}37.2423 (0.6724)/\\ 58.0833 (1.9443)\end{tabular}}          \\ \hline
GraphAT                     & \multicolumn{2}{c}{\begin{tabular}[c]{@{}c@{}}24.1835 (2.5629)/\\ 36.5068 (2.8531)\end{tabular}}          & \multicolumn{2}{c}{\begin{tabular}[c]{@{}c@{}}26.5689 (1.7934)/\\ 40.2971 (2.2835)\end{tabular}}          & \multicolumn{2}{c}{\begin{tabular}[c]{@{}c@{}}27.0025 (1.8779)/\\ 41.4157 (2.4305)\end{tabular}}          & \multicolumn{2}{c}{\begin{tabular}[c]{@{}c@{}}26.3760 (1.4645)/\\ 40.5200 (1.9659)\end{tabular}}          & \multicolumn{2}{c}{\begin{tabular}[c]{@{}c@{}}27.0758 (1.8094)/\\ 41.8067 (2.3826)\end{tabular}}          & \multicolumn{2}{c}{\begin{tabular}[c]{@{}c@{}}29.2397 (3.3257)/\\ 44.8820 (3.7756)\end{tabular}}          \\ \hline
AT-TNDS                     & \multicolumn{2}{c}{\begin{tabular}[c]{@{}c@{}}24.7358 (2.5316)/\\ 37.3832 (3.1755)\end{tabular}}          & \multicolumn{2}{c}{\begin{tabular}[c]{@{}c@{}}27.7863 (2.3676)/\\ 42.1406 (3.2536)\end{tabular}}          & \multicolumn{2}{c}{\begin{tabular}[c]{@{}c@{}}28.1592 (2.3334)/\\ 43.1551 (3.0992)\end{tabular}}          & \multicolumn{2}{c}{\begin{tabular}[c]{@{}c@{}}27.7167 (2.2004)/\\ 42.4776 (2.9184)\end{tabular}}          & \multicolumn{2}{c}{\begin{tabular}[c]{@{}c@{}}28.3284 (2.2946)/\\ 43.6221 (2.9969)\end{tabular}}          & \multicolumn{2}{c}{\begin{tabular}[c]{@{}c@{}}27.5406 (1.5208)/\\ \textbf{42.4567 (1.8109)}\end{tabular}}          \\ \hline
Ours                        & \multicolumn{2}{c}{\begin{tabular}[c]{@{}c@{}}24.4491 (1.7130)/\\ 36.8813 (2.1736)\end{tabular}} & \multicolumn{2}{c}{\textbf{\begin{tabular}[c]{@{}c@{}}25.6935 (0.9964)/\\ 39.1491 (0.8991)\end{tabular}}} & \multicolumn{2}{c}{\textbf{\begin{tabular}[c]{@{}c@{}}26.0884 (0.9818)/\\ 39.9803 (1.0121)\end{tabular}}} & \multicolumn{2}{c}{\textbf{\begin{tabular}[c]{@{}c@{}}25.8660 (0.9360)/\\ 39.5416 (0.7852)\end{tabular}}} & \multicolumn{2}{c}{\textbf{\begin{tabular}[c]{@{}c@{}}26.2960 (0.8723)/\\ 40.4473 (0.8074)\end{tabular}}} & \multicolumn{2}{c}{\begin{tabular}[c]{@{}c@{}} \textbf{27.5366 (0.9962)}/\\ 42.5309 (1.0727)\end{tabular}} \\ \hline
\end{tabular}}
\end{table*}

We further examines the robustness of our traffic forecasting model against adversarial attacks, in comparison to five baselines, under four different attack strengths ($\lambda ={40,60,80,100}$). The results displayed in Figure~\ref{fig:PGD_Random} demonstrate that our method demonstrates superior performance across all attack strengths, exemplified by a 13.9842\% and 2.8602\% improvement on PeMS-BAY and PeMS-D4 under 100\% attack strength. Notably, TRADE method performed inferiorly to other adversarial training methods (AT, Mixup, GraphAT, AT-TNDS) under stronger attack strengths, likely due to the trade-off between clean performance and adversarial performance. Furthermore, the results indicate that AT-TNDS outperforms almost all other baselines, which verifies the effectiveness of dynamically selecting a subset of nodes as the adversarial nodes. However, AT-TNDS performed worse than our method, as it encountered instability during training and was unable to handle dynamic adversarial traffic attacks.

\subsection{EQ2: Ablation Study}
\begin{figure}[tb]
\centering
\subfigure[MAE]{
\begin{minipage}[]{0.5\linewidth}
\centering
\includegraphics[width=1.5in]{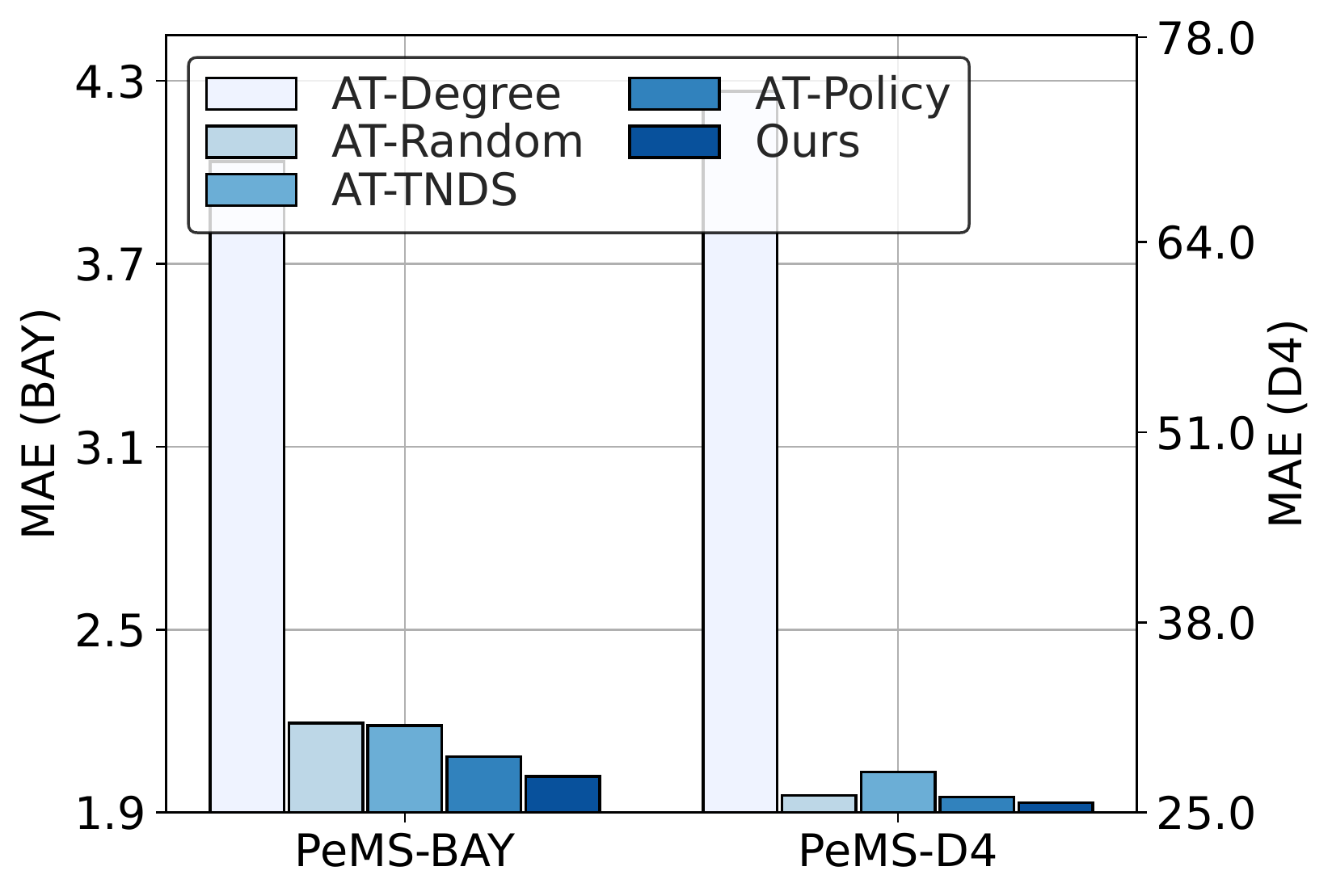}
\end{minipage}%
}%
\subfigure[STD]{
\begin{minipage}[]{0.5\linewidth}
\centering
\includegraphics[width=1.5in]{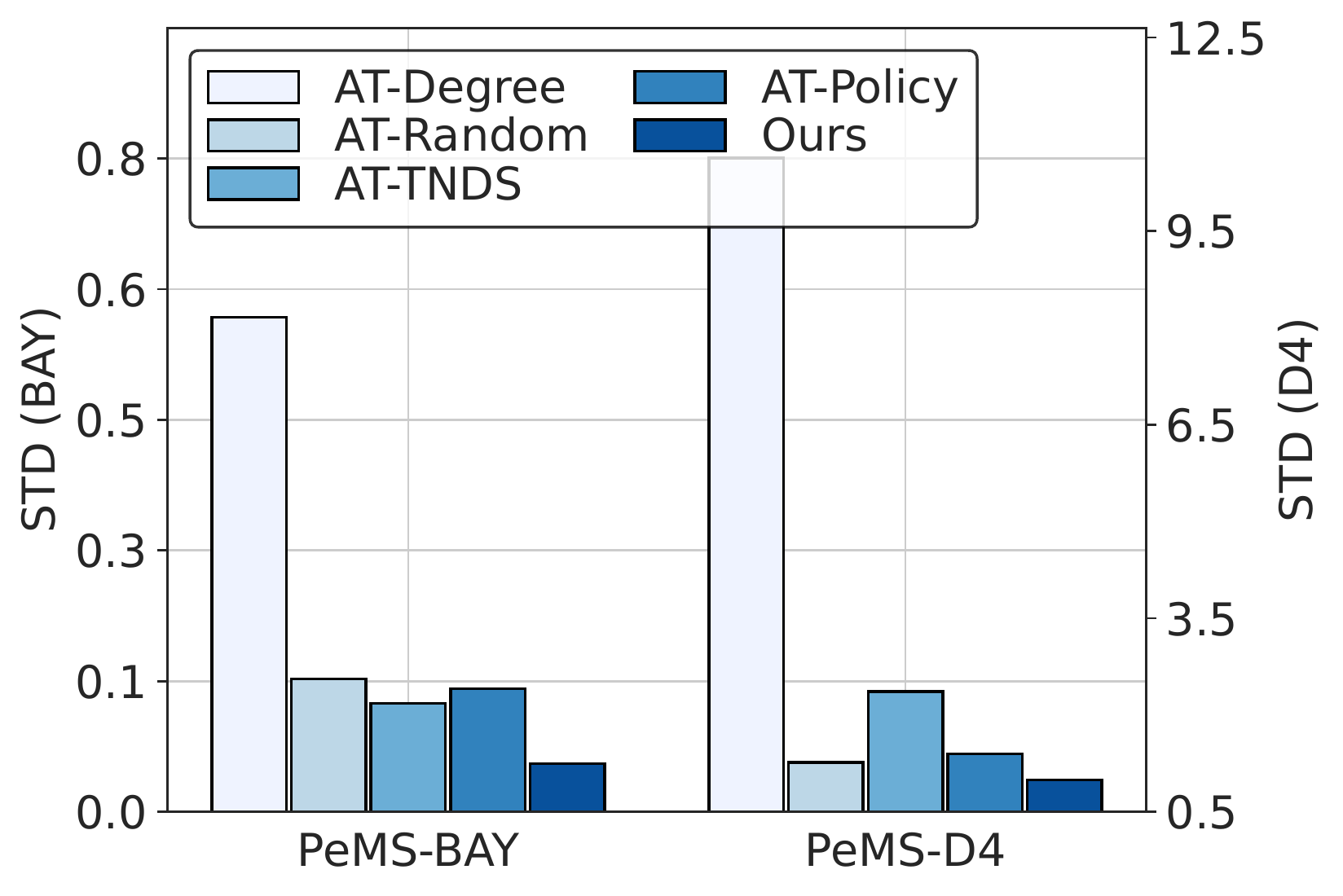}
\end{minipage}%
}%
\centering
\setlength{\belowcaptionskip}{-0.6cm}
\caption{ Ablation studies on two datasets. "BAY" and "D4" denote PeMS-BAY and PeMS-D4, respectively.}\label{fig:ablation_study}
\end{figure}

In order to answer EQ 2, we examined the impact of different components of our adversarial training framework on the performance of traffic forecasting models by conducting an ablation study using the mean absolute error (MAE) metric on the PeMS-BAY dataset. We evaluated four variations of our method: (1) AT-Degree, which selects nodes based on their normalized degree in a static manner, (2) AT-Random, which selects nodes randomly in a dynamic manner, (3) AT-TNDS, which selects nodes based on a spatiotemporal-dependent method, (4) AT-Policy, which uses a pre-trained policy network to choose nodes without self-distillation, and (5) our method, which uses a pre-trained policy network with self-distillation regularization to choose nodes. As reported in Figure~\ref{fig:ablation_study}, removing any component causes significant performance degradation. In particular, we observed a significant degradation when using the static node selection strategy, which demonstrates the effectiveness of our adversarial training framework. Secondly, by comparing AT-Random and AT-Policy, we found that the policy network plays an important role in selecting the subset of nodes as adversarial examples. Lastly, we also observed that self-distillation significantly improves the stability of adversarial training; for example, by removing self-distillation, the standard deviation was reduced by 60\% and 27.0775\% on PeMS-BAY and PeMS-D4, respectively.

\eat{

\begin{table}[ht]
\centering
\caption{Ablation study}\label{tab:ablation_study}
\scalebox{0.6}{\begin{tabular}{ccccccc}
\hline
          & Dynamic      & Spatiotemporal-dependent & Learning-based & Regularization & MAE & STD \\ \hline
AT-degree &              &                          &                &                &     &     \\
AT-Random & $\checkmark$ &                          &                &                &     &     \\
AT-TNDS   & $\checkmark$ & $\checkmark$             &                &                &     &     \\
AT-Policy & $\checkmark$ & $\checkmark$             & $\checkmark$   &                &     &     \\
Our       & $\checkmark$ & $\checkmark$             & $\checkmark$   & $\checkmark$   &     &     \\ \hline
\end{tabular}}
\end{table}}

\begin{figure}[thb]
\centering
\subfigure[Inner iteration b]{
\begin{minipage}[]{0.5\linewidth}
\centering
\includegraphics[width=1.4in]{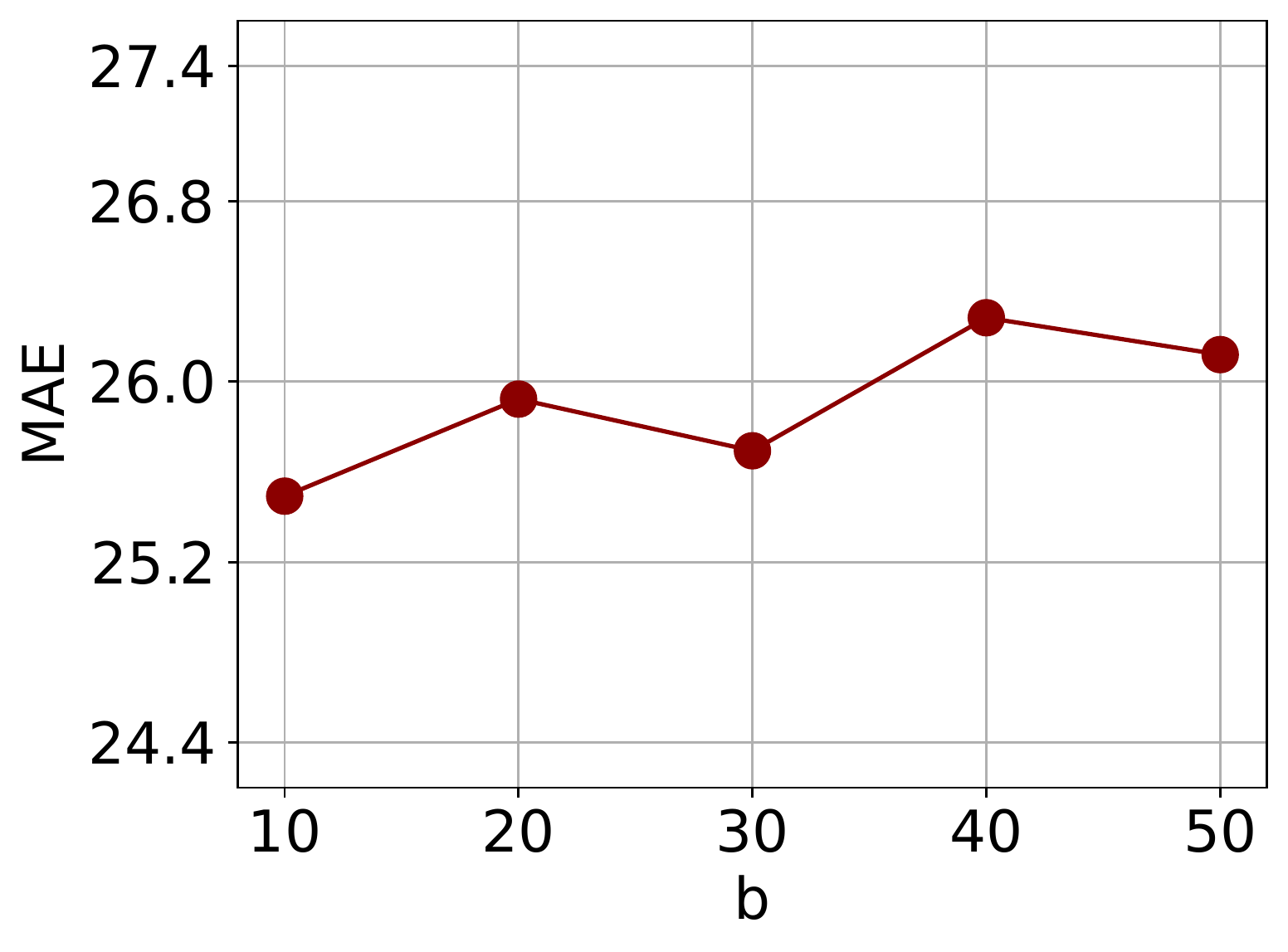}
\end{minipage}%
}%
\subfigure[Regularization parameter $\alpha$]{
\begin{minipage}[]{0.5\linewidth}
\centering
\includegraphics[width=1.4in]{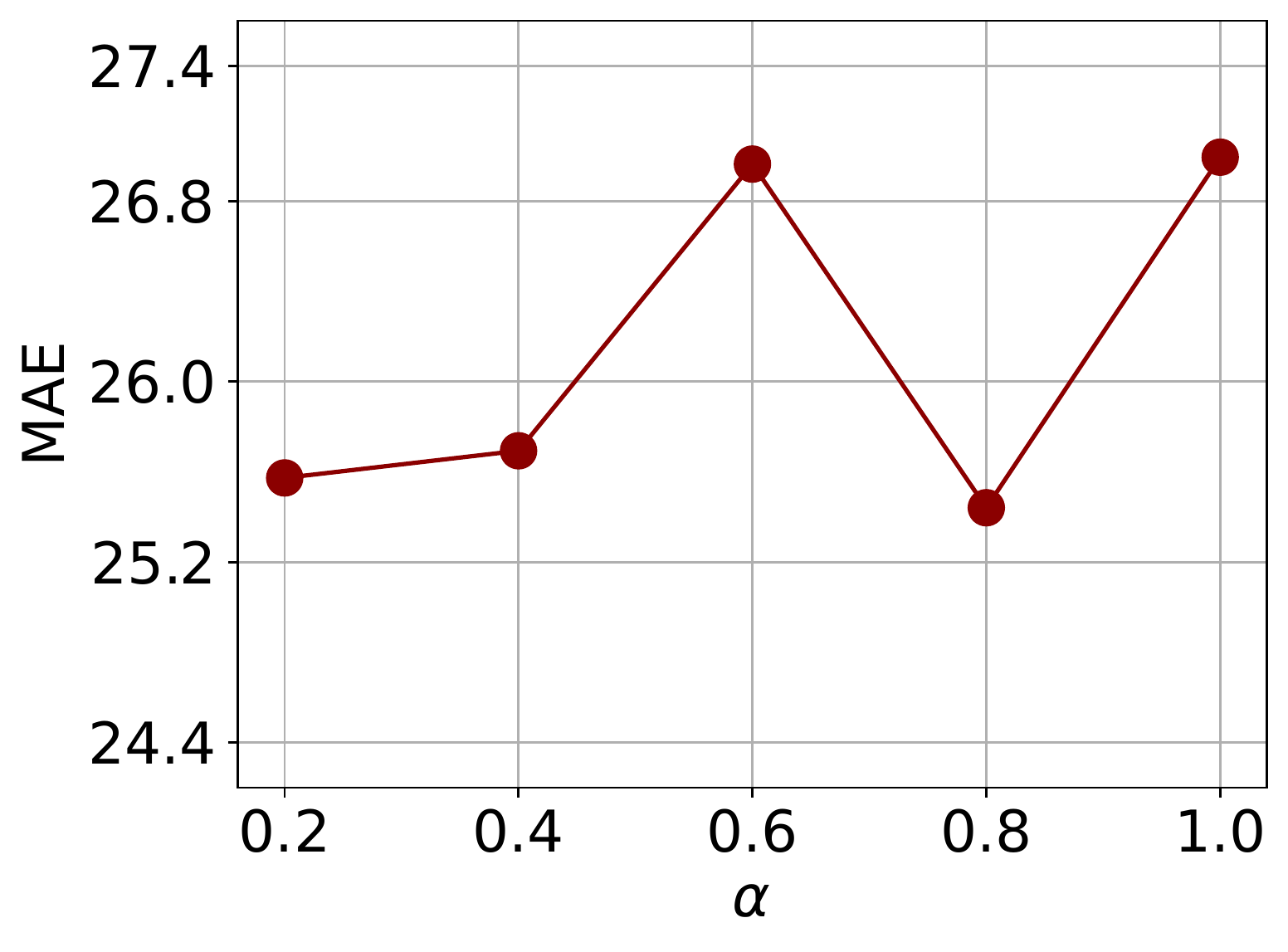}
\end{minipage}%
}%
\centering
\setlength{\belowcaptionskip}{-0.5cm}
\caption{ Effect on different parameters.}\label{fig:para}
\end{figure}

\subsection{EQ3: Parameter Analysis}
To answer EQ 3, we conducted a sensitivity analysis to evaluate the impact of hyperparameters on the performance of our adversarial training framework using the PeMS-D4 dataset as an example. The parameters studied were the number of inner iteration (b)  and the regularization parameter ($\alpha$), while all other parameters remained constant. The results showed an overall increasing trend in performance with increasing number of inner iterations, with a peak at 30 iterations (Figure~\ref{fig:para} (a)). The performance under different regularization parameters showed an initial increase, reaching the lowest point at $\alpha$ = 0.6, then decreasing at $\alpha$ = 0.8, before increasing again at the highest value (Figure~\ref{fig:para} (b)).

\subsection{Case Study}
\begin{figure}[thb]
\centering
\subfigure[Non-defense]{
\begin{minipage}[]{0.45\textwidth}
\centering
\includegraphics[width=0.7\textwidth]{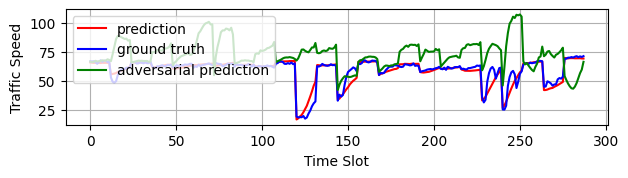}
\end{minipage}%
}%

\subfigure[Defense by adversarial training]{
\begin{minipage}[]{0.45\textwidth}
\centering
\includegraphics[width=0.7\textwidth]{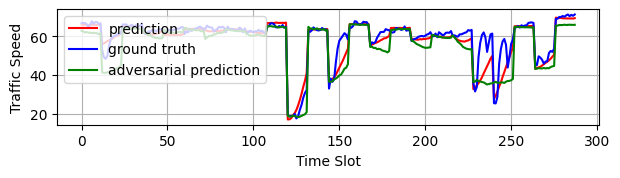}
\end{minipage}%
}%
\centering
\setlength{\belowcaptionskip}{-0.5cm}
\caption{ Comparison of traffic forecasting model performance under adversarial attack. (a) Results without defense, showing biased predictions under attack. (b) Results with adversarial training, showing improved robustness and similar predictions to original model. }\label{fig:case_study}
\end{figure}

In this section, we conduct the case study to show the effectiveness of our adversarial training framework. The case study presented in the Figure~\ref{fig:case_study} provides a visual representation of the effectiveness of our proposed adversarial training framework for spatiotemporal traffic forecasting tasks. Figure~\ref{fig:case_study} (a) illustrates the results of a traffic forecasting model that has not been defended against adversarrial attacks, while Figure~~\ref{fig:case_study} (b) illustrates the results of the same model with our proposed adversarial training defense. It is clear from the figures that the model without defense provides biased predictions under adversarial attack, while the model with defense (adversarial training) maintains its prediction accuracy and is able to provide similar results as the original predictions.

\section{Related work}
In this section, we briefly introduce the related topics including spatiotemporal traffic forecasting and adversarial training.
\subsection{Spatiotemporal Traffic Forecasting}
In recent years, deep learning has found widespread applications in diverse domains, including job skill valuation~\cite{sun2021market}, time series prediction~\cite{zhou2015smiler} and spatiotemporal traffic forecasting~\cite{zhang2020semi}.
Among these applications, spatiotemporal traffic forecasting plays a crucial role in the success of intelligent transportation systems~\cite{LeiMSW22, Fang0ZHCGJ22, HanDSFL021, WuGCXVMY21, LiZBLLZ20, YeSDFTX19, PanLW00Z19}. The ability to predict traffic patterns in both space and time is critical for effective traffic management and improved travel experiences for commuters. To address these forecasting problems, deep learning models have been extensively explored due to their superior ability to model the complex spatial and temporal dependencies present in traffic data. Various methods have been proposed to enhance the accuracy of traffic forecasting, such as STGCN~\cite{STGCN}, DCRNN~\cite{DCRNN} and GraphWave Net~\cite{Graph_Wave_Net}, each of which utilizes different techniques to capture spatial and temporal information. Despite the advancements made, the vulnerability and adversarial robustness of spatiotemporal traffic forecasting models remain an unexplored area of research.
\subsection{Adversarial Training}
The existing literature has explored various methods to improve the robustness of deep learning models against adversarial attacks~\cite{madry2018towards-PGD, AvdiukhinMYZ19, JiaLZKK19, KJW021, ShiZ022, PengLZLYL022}. One popular approach is adversarial training, which aims to enhance the intrinsic robustness of the model by augmenting the training data with adversarial examples. This approach is based on a min-max optimization problem, where the solution is found by identifying the worst-case optimum. Adversarial training methods have been proposed for different types of data, such as images and graphs. For example, TRADE~\cite{tramer2019adversarial-TRADE} is a method that balances the trade-off between clean performance and robustness, while GraphAT~\cite{FengHTC21} is a variant of adversarial training specifically designed for graph-structured data. However, current adversarial training methods mostly focus on the static domain, such as images and graphs, and the dynamic domain, such as spatiotemporal forecasting, is less explored.

\section{Conclusion}\label{sec:con}
In conclusion, this paper presents a novel framework for incorporating adversarial training into spatiotemporal traffic forecasting tasks. We reveal that traditional adversarial training methods in static domains are not suitable for defending against dynamic adversarial attacks in traffic forecasting tasks. To this end, we propose a reinforcement learning-based method to learn the optimal strategy for selecting adversarial examples, which improves defense against dynamic attacks and reduces overfitting. Additionally, we introduce a self-knowledge distillation regularization to overcome the "forgetting issue" caused by constantly changing adversarial nodes during training. We evaluate our approach on two real-world traffic datasets and demonstrate its superiority over other baselines. Our method effectively enhances the adversarial robustness of spatiotemporal traffic forecasting models, which is essential in ITS to provide accurate predictions and avoid negative consequences such as congestion and delays. \eat{This research provides a new method to prevent the adversarial attack in the ITS area, and it will benefit the development of ITS and intelligent traffic control. The future work include the scalability and robustness test on different datasets and different attacks.}

\begin{acks}
This research was supported in part by the National Natural Science Foundation of China under Grant No.62102110, Guangzhou Science and Technology Plan Guangzhou-HKUST(GZ) Joint Project No. 2023A03J0144, and Foshan HKUST Projects (FSUST21-FYTRI01A, FSUST21-FYTRI02A).
\end{acks}

\bibliographystyle{ACM-Reference-Format}
\balance
\bibliography{sample-base}

\appendix

\section{Spatiotemporal Adversarial Training}~\label{sec:appdix-alg}
The training process is divided into two stages. In the first stage, we train the policy network using Algorithm~\ref{alg:policy_network_learning}. In the second stage, we use the pre-trained policy network to select the adversarial nodes for computational efficiency, and then compute the adversarial examples by using the PGD method. The adversarial examples are computed using Equation~\ref{eq:white-box_ST-PGD} with the adversarial training loss $\mathcal{L}_{AT}$ in Equation~\ref{eq:AT_loss}. Finally, we update the forecasting model parameters $\theta$ using the Adam optimizer in Algorithm~\ref{alg:adversarial_learning}.

\begin{algorithm}[!]
\caption{Spatiotemporal Traffic Adversarial Training}\label{alg:adversarial_learning}
\KwIn{
 Traffic data, pre-trained policy network, number of epochs E, maximum perturbation budget $\epsilon$, adversarial node budget $\eta$.
}
\KwResult{ Spatiotemporal traffic forecasting model $\mathcal{F}_{\theta}(\cdot)$}
Initialize spatiotemporal traffic forecasting parameter $\theta $\;
{\For { $i=1$  to $E$ }
{
Obtain the solution  $\Omega^{(p)}$ by Equation~\ref{eq:Solution} based on pre-trained policy model \;
Compute adversarial node indicator based on the solution  $\Omega^{(p)}$  \;
Generate perturbed adversarial features $\mathbf{{X}'}_{t-\tau:t}$ by Equation~\ref{eq:white-box_ST-PGD}\;  
Compute the adversarial training loss by Equation~\ref{eq:AT_loss}\;
 Update $\theta$ using Adam optimization with the gradients of $\mathcal{L}_{AT}$ \;
}}
Return $\mathcal{F}_{\theta}(\cdot)$.
\end{algorithm}

\section{Supplementary Experiments}

\subsection{Implements Details}
We conduct our experiments using Pytorch on a Linux Centos Server with 12 RTX 3090 GPUs and 2 RTX A40 GPUs. The traffic data is normalized to the range [0,1], and the input and output lengths are set to $\tau=12$ and $T=12$, respectively.  We adopt the attack setting as outlined in~\cite{liu2022practical}, utilizing PGD-Random, PGD-PR, PGD-Centrality, PGD-Degree, and PGD-TNDS as the attackers. The attack steps are set to 5, with a step size of $\gamma=0.1$. To thoroughly assess the adversarial robustness of our forecasting models, the attacks are conducted in a white-box setting, following the methodology in~\cite{liu2022practical}.  The regularization parameter $\alpha$ is set to $0.4$. The perturbation magnitude $\epsilon$ is 0.5 for both training and testing. During training, we select 10\% of the total nodes as adversarial examples at each epoch, while in testing, we use a stronger attack strength and select 20\% of the total nodes as adversarial examples. We conducted the experiments five times and present the average results along with the standard deviations (STD) of the metrics.

\subsection{Baselines}
Adversarial training is an effective approach to improving the robustness of machine learning models against adversarial attacks. In this work, we evaluate the performance of several state-of-the-art adversarial training methods including:
1) AT~\cite{madry2018towards-PGD}: This method applies traditional adversarial training by selecting adversarial nodes randomly.
2) Mixup~\cite{zhang2018mixup}: A data augmentation technique that blends pairs of training examples and their corresponding labels in a weighted combination to enhance the robustness of the model.
3) TRADE~\cite{tramer2019adversarial-TRADE}: This method utilizes a trade-off based approach to perform adversarial training. It balances the accuracy of the model and its robustness against adversarial attacks.
4) GraphAT~\cite{FengHTC21}: This method is specifically designed for graph-structured data and applies adversarial training to graph-based models.
5) AT-TNDS~\cite{liu2022practical}: This method implements a dynamic strategy for selecting the victim nodes and performs traditional adversarial training. The dynamic selection strategy allows for adaptability to changing adversarial patterns during training.

\subsection{Adversarial Robustness Comparison}\label{sec:ARC}
In order to further evaluate the effectiveness of our proposed method (RDAT), we conduct adversarial training experiments on other methods including ASTGCN~\cite{ASTGCN} and STGCN~\cite{STGCN} on PeMS-D4. The results of these experiments are presented in Table~\ref{tab:main_results_astgcn_PeMS-D4} and \ref{tab:main_results_stgcn_PeMS-D4}. We observed that adversarial training shows a considerable amount of variability in the ASTGCN and STGCN models. This might be associated with the model architecture, as Graphwave Net has the capability to directly learn the spatial relationships through the data, potentially leading to assigning smaller weights to nodes that are more susceptible to attacks.

\begin{table*}[thb]
\caption{Adversarial robustness performance ( MAE (std) / RMSE (std)) on dataset PEMS-D4 on ASTGCN}\label{tab:main_results_astgcn_PeMS-D4}
\scalebox{0.9}{
\begin{tabular}{c|clclclclclcl}
\hline
Method & \multicolumn{2}{c}{Non-attack}                                                                            & \multicolumn{2}{c}{PGD-Random}                                                                            & \multicolumn{2}{c}{PGD-PR}                                                                               & \multicolumn{2}{c}{PGD-Centrality}                                                                        & \multicolumn{2}{c}{PGD-Degree}                                                                            & \multicolumn{2}{c}{PGD-TNDS}                                                                               \\ \hline
Non-defense                 & \multicolumn{2}{c}{\begin{tabular}[c]{@{}c@{}}27.2793 (1.1845)/\\ 40.0507 (0.9279)\end{tabular}}          & \multicolumn{2}{c}{\begin{tabular}[c]{@{}c@{}}111.8937 (5.5077)/\\ 212.5077 (15.2161)\end{tabular}}       & \multicolumn{2}{c}{\begin{tabular}[c]{@{}c@{}}113.0682 (4.8666)\\ 213.7147 (13.1761)\end{tabular}}       & \multicolumn{2}{c}{\begin{tabular}[c]{@{}c@{}}111.4333 (7.1745)\\ 212.2865 (19.4604)/\end{tabular}}       & \multicolumn{2}{c}{\begin{tabular}[c]{@{}c@{}}112.6771 (5.9415)/\\ 214.8724 (17.1004)\end{tabular}}       & \multicolumn{2}{c}{\begin{tabular}[c]{@{}c@{}}120.1700 (14.057)/\\ 221.2135 (13.4)\end{tabular}}           \\ \hline
AT                          & \multicolumn{2}{c}{\begin{tabular}[c]{@{}c@{}}\textbf{34.6683 (1.7694)}/\\ 50.0380 (3.4274)\end{tabular}} & \multicolumn{2}{c}{\begin{tabular}[c]{@{}c@{}}60.9065 (3.6145)/\\ 93.0503 (6.6562)\end{tabular}}          & \multicolumn{2}{c}{\begin{tabular}[c]{@{}c@{}}63.6593 (3.3482)\\ 96.4130 (6.5096)\end{tabular}}          & \multicolumn{2}{c}{\begin{tabular}[c]{@{}c@{}}59.8317 (4.5009)/\\ 93.5280 (9.1634)\end{tabular}}          & \multicolumn{2}{c}{\begin{tabular}[c]{@{}c@{}}65.4675 (3.4941)/\\ 100.2619 (6.6652)\end{tabular}}         & \multicolumn{2}{c}{\begin{tabular}[c]{@{}c@{}}\textbf{69.5971 (3.5078)}/\\ \textbf{ 107.7790 (5.7359)}\end{tabular}}          \\ \hline
Mixup                       & \multicolumn{2}{c}{\begin{tabular}[c]{@{}c@{}}38.3808 (7.3672)/\\ 54.2371 (9.8325)\end{tabular}}          & \multicolumn{2}{c}{\begin{tabular}[c]{@{}c@{}}61.5212 (11.9578)/\\ 90.3154 (19.0033)\end{tabular}}        & \multicolumn{2}{c}{\begin{tabular}[c]{@{}c@{}}62.5962 (11.9046)\\ 91.6064 (18.9730)\end{tabular}}        & \multicolumn{2}{c}{\begin{tabular}[c]{@{}c@{}}60.3004 (14.2821)/\\ 90.7889 (22.9789)\end{tabular}}        & \multicolumn{2}{c}{\begin{tabular}[c]{@{}c@{}}63.6662 (13.1854)/\\ 94.3743 (20.8728)\end{tabular}}        & \multicolumn{2}{c}{\begin{tabular}[c]{@{}c@{}}75.9225 (15.3793)/\\ 113.1050 (23.9088)\end{tabular}}        \\ \hline
TRADE                       & \multicolumn{2}{c}{\begin{tabular}[c]{@{}c@{}}36.8796 (4.1128)/\\ 52.2104 (4.5704)\end{tabular}} & \multicolumn{2}{c}{\begin{tabular}[c]{@{}c@{}}68.2222 (5.6103)/\\ 111.0874 (10.0838)\end{tabular}}        & \multicolumn{2}{c}{\begin{tabular}[c]{@{}c@{}}70.5183 (5.3248)\\ 113.8450 (11.0367)\end{tabular}}        & \multicolumn{2}{c}{\begin{tabular}[c]{@{}c@{}}66.6756 (4.9103)/\\ 110.7533 (9.6369)\end{tabular}}         & \multicolumn{2}{c}{\begin{tabular}[c]{@{}c@{}}71.5576 (3.9394)/\\ 116.7629 (8.5226)\end{tabular}}         & \multicolumn{2}{c}{\begin{tabular}[c]{@{}c@{}}81.2129 (5.4850)/\\ 131.8080 (7.6451)\end{tabular}}          \\ \hline
GraphAT                     & \multicolumn{2}{c}{\begin{tabular}[c]{@{}c@{}}37.1787 (3.2990)/\\ 53.3633 (5.0211)\end{tabular}}          & \multicolumn{2}{c}{\begin{tabular}[c]{@{}c@{}}63.6767 (4.8531)/\\ 95.3386 (7.6426)\end{tabular}}          & \multicolumn{2}{c}{\begin{tabular}[c]{@{}c@{}}67.3017 (5.0861)\\ 100.1515 (7.8135)\end{tabular}}         & \multicolumn{2}{c}{\begin{tabular}[c]{@{}c@{}}63.1912 (5.3288)/\\ 96.9107 (8.9905)\end{tabular}}          & \multicolumn{2}{c}{\begin{tabular}[c]{@{}c@{}}69.1714 (5.1078)/\\ 104.1556 (7.6357)\end{tabular}}         & \multicolumn{2}{c}{\begin{tabular}[c]{@{}c@{}}74.6318 (3.0041)/\\ 114.2343 (5.2069)\end{tabular}}          \\ \hline
AT-TNDS                     & \multicolumn{2}{c}{\begin{tabular}[c]{@{}c@{}}29.5898 (2.6930)/\\ 42.7446 (3.0345)\end{tabular}}          & \multicolumn{2}{c}{\begin{tabular}[c]{@{}c@{}}66.0547 (2.6993)/\\ 99.2921 (7.0982)\end{tabular}}          & \multicolumn{2}{c}{\begin{tabular}[c]{@{}c@{}}67.9138 (2.2889)\\ 101.9280 (6.1251)\end{tabular}}         & \multicolumn{2}{c}{\begin{tabular}[c]{@{}c@{}}65.5843 (4.1708)/\\ 100.1370 (10.0935)\end{tabular}}        & \multicolumn{2}{c}{\begin{tabular}[c]{@{}c@{}}68.8698 (3.1337)/\\ 104.3428 (7.4617)\end{tabular}}         & \multicolumn{2}{c}{\begin{tabular}[c]{@{}c@{}}71.6321 (3.7687)/\\ 108.2953 (9.6008)\end{tabular}}          \\ \hline
Ours                        & \multicolumn{2}{c}{\begin{tabular}[c]{@{}c@{}}35.0078 (3.9493)/\\ \textbf{49.1218 (4.8374)}\end{tabular}} & \multicolumn{2}{c}{\textbf{\begin{tabular}[c]{@{}c@{}}57.1716 (4.4675)/\\ 83.5783 (7.3353)\end{tabular}}} & \multicolumn{2}{c}{\textbf{\begin{tabular}[c]{@{}c@{}}59.3260 (4.8961)\\ 86.5329 (7.7137)\end{tabular}}} & \multicolumn{2}{c}{\textbf{\begin{tabular}[c]{@{}c@{}}56.1105 (4.5573)/\\ 83.6596 (8.0753)\end{tabular}}} & \multicolumn{2}{c}{\textbf{\begin{tabular}[c]{@{}c@{}}61.4775 (4.1038)/\\ 90.8641 (6.6173)\end{tabular}}} & \multicolumn{2}{c}{\begin{tabular}[c]{@{}c@{}}71.8013 (4.9489)/\\ 111.8190 (9.0837)\end{tabular}} \\ \hline
\end{tabular}}
\end{table*}

\begin{table*}[thb]
\caption{Adversarial robustness performance ( MAE (std) / RMSE (std)) on dataset PEMS-D4 on STGCN}\label{tab:main_results_stgcn_PeMS-D4}
\scalebox{0.9}{
\begin{tabular}{c|clclclclclcl}
\hline
Method & \multicolumn{2}{c}{Non-attack}                                                                     & \multicolumn{2}{c}{PGD-Random}                                                                            & \multicolumn{2}{c}{PGD-PR}                                                                                  & \multicolumn{2}{c}{PGD-Centrality}                                                                          & \multicolumn{2}{c}{PGD-Degree}                                                                              & \multicolumn{2}{c}{PGD-TNDS}                                                                                \\ \hline
Non-defense                 & \multicolumn{2}{c}{\begin{tabular}[c]{@{}c@{}}27.8656 (0.5019)/\\ 41.6828 (0.4553)\end{tabular}}   & \multicolumn{2}{c}{\begin{tabular}[c]{@{}c@{}}88.6441 (4.1227)/\\ 128.5004 (6.7317)\end{tabular}}         & \multicolumn{2}{c}{\begin{tabular}[c]{@{}c@{}}94.3075 (4.8237)/\\ 144.2537 (9.1144)\end{tabular}}           & \multicolumn{2}{c}{\begin{tabular}[c]{@{}c@{}}86.0901 (7.3620)/\\ 154.8890 (18.2470)\end{tabular}}          & \multicolumn{2}{c}{\begin{tabular}[c]{@{}c@{}}94.8585 (6.0286)/\\ 155.8989 (11.8078)\end{tabular}}          & \multicolumn{2}{c}{\begin{tabular}[c]{@{}c@{}}86.0901 (7.3620)/\\ 154.8890 (18.2470)\end{tabular}}          \\ \hline
AT                          & \multicolumn{2}{c}{\begin{tabular}[c]{@{}c@{}}33.3386 (3.9365)/\\ \textbf{47.1510 (3.5193)}\end{tabular}}   & \multicolumn{2}{c}{\textbf{\begin{tabular}[c]{@{}c@{}}39.8153 (5.5781)/\\ 56.2785 (7.2508)\end{tabular}}} & \multicolumn{2}{c}{\begin{tabular}[c]{@{}c@{}}44.2567 (7.8589)/\\ 63.2609 (12.7096)\end{tabular}}           & \multicolumn{2}{c}{\begin{tabular}[c]{@{}c@{}}48.7779 (10.4840)/\\ 75.9330 (20.7510)\end{tabular}}          & \multicolumn{2}{c}{\begin{tabular}[c]{@{}c@{}}48.3946 (10.5063)/\\ 71.7793 (18.3587)\end{tabular}}          & \multicolumn{2}{c}{\begin{tabular}[c]{@{}c@{}}50.6256 (11.2085)/\\ 77.1303 (20.7762)\end{tabular}}          \\ \hline
Mixup                       & \multicolumn{2}{c}{\begin{tabular}[c]{@{}c@{}}38.2206 (2.7247)/\\ 52.9308 (3.5875)\end{tabular}}   & \multicolumn{2}{c}{\begin{tabular}[c]{@{}c@{}}46.7455 (4.3982)/\\ 64.0452 (5.2427)\end{tabular}}          & \multicolumn{2}{c}{\begin{tabular}[c]{@{}c@{}}48.2348 (5.2173)/\\ 65.9415 (6.4053)\end{tabular}}            & \multicolumn{2}{c}{\begin{tabular}[c]{@{}c@{}}50.7168 (5.3034)/\\ 73.2369 (10.1490)\end{tabular}}           & \multicolumn{2}{c}{\begin{tabular}[c]{@{}c@{}}49.1789 (5.1138)/\\ 68.4766 (7.6452)\end{tabular}}            & \multicolumn{2}{c}{\begin{tabular}[c]{@{}c@{}}50.8233 (7.3831)/\\ 70.6203 (11.0908)\end{tabular}}           \\ \hline
TRADE                       & \multicolumn{2}{c}{\begin{tabular}[c]{@{}c@{}}34.5135 (4.7824)/\\ 49.4893 (5.7589)\end{tabular}}   & \multicolumn{2}{c}{\begin{tabular}[c]{@{}c@{}}58.1406 (14.6082)/\\ 82.7603 (19.7603)\end{tabular}}        & \multicolumn{2}{c}{\begin{tabular}[c]{@{}c@{}}65.8004 (18.4610)/\\ 96.5914 (28.5557)\end{tabular}}          & \multicolumn{2}{c}{\begin{tabular}[c]{@{}c@{}}69.3065 (15.2959)/\\ 115.3199 (28.3532)\end{tabular}}         & \multicolumn{2}{c}{\begin{tabular}[c]{@{}c@{}}70.8713 (18.3982)/\\ 110.2185 (32.3967)\end{tabular}}         & \multicolumn{2}{c}{\begin{tabular}[c]{@{}c@{}}76.3904 (20.9501)/\\ 123.2154 (38.2284)\end{tabular}}         \\ \hline
GraphAT                     & \multicolumn{2}{c}{\begin{tabular}[c]{@{}c@{}}47.1434 (12.6269)/\\ 60.8673 (12.0997)\end{tabular}} & \multicolumn{2}{c}{\begin{tabular}[c]{@{}c@{}}55.8650 (12.1506)/\\ 72.7158 (12.2657)\end{tabular}}        & \multicolumn{2}{c}{\begin{tabular}[c]{@{}c@{}}59.2201 (12.6541)/\\ 78.5934 (15.0033)\end{tabular}}          & \multicolumn{2}{c}{\begin{tabular}[c]{@{}c@{}}61.9119 (12.6672)/\\ 86.6466 (17.7716)\end{tabular}}          & \multicolumn{2}{c}{\begin{tabular}[c]{@{}c@{}}62.5326 (14.0952)/\\ 85.1818 (18.9048)\end{tabular}}          & \multicolumn{2}{c}{\begin{tabular}[c]{@{}c@{}}62.5210 (12.8110)/\\ 85.6943 (16.0209)\end{tabular}}          \\ \hline
AT-TNDS                     & \multicolumn{2}{c}{\begin{tabular}[c]{@{}c@{}}\textbf{31.8291 (1.3078)}/\\ 66.6957 (1.2478)\end{tabular}}   & \multicolumn{2}{c}{\begin{tabular}[c]{@{}c@{}}46.8909 (1.1684)/\\ 66.6957 (1.2478)\end{tabular}}          & \multicolumn{2}{c}{\begin{tabular}[c]{@{}c@{}}49.4102 (1.8226)/\\ 70.5907 (1.2680)\end{tabular}}            & \multicolumn{2}{c}{\begin{tabular}[c]{@{}c@{}}55.1733 (2.1787)/\\ 91.3877 (6.6509)\end{tabular}}            & \multicolumn{2}{c}{\begin{tabular}[c]{@{}c@{}}53.2867 (1.4795)/\\ 80.0251 (1.1357)\end{tabular}}            & \multicolumn{2}{c}{\begin{tabular}[c]{@{}c@{}}54.3390 (2.4273)/\\ 83.5821 (6.1762)\end{tabular}}            \\ \hline
Ours                        & \multicolumn{2}{c}{\begin{tabular}[c]{@{}c@{}}34.2860 (5.9462)/\\ 48.8814 (6.9747)\end{tabular}}   & \multicolumn{2}{c}{\begin{tabular}[c]{@{}c@{}}41.5189 (10.2116)/\\ 58.0855 (12.5508)\end{tabular}}        & \multicolumn{2}{c}{\textbf{\begin{tabular}[c]{@{}c@{}}43.7104 (11.3853)/\\ 61.4063 (15.0858)\end{tabular}}} & \multicolumn{2}{c}{\textbf{\begin{tabular}[c]{@{}c@{}}46.0493 (11.5675)/\\ 68.5447 (18.2039)\end{tabular}}} & \multicolumn{2}{c}{\textbf{\begin{tabular}[c]{@{}c@{}}45.4336 (11.7635)/\\ 65.1032 (16.5949)\end{tabular}}} & \multicolumn{2}{c}{\textbf{\begin{tabular}[c]{@{}c@{}}44.9669 (9.8571)/\\ 64.15829 (13.1911)\end{tabular}}} \\ \hline
\end{tabular}}
\end{table*}

\subsection{Adversarial Robustness under Various Attack Intensities }
\begin{figure*}[thb]
\centering
\subfigure[PGD-PR]{
\begin{minipage}[]{0.25\linewidth}
\centering
\includegraphics[width=1.2in]{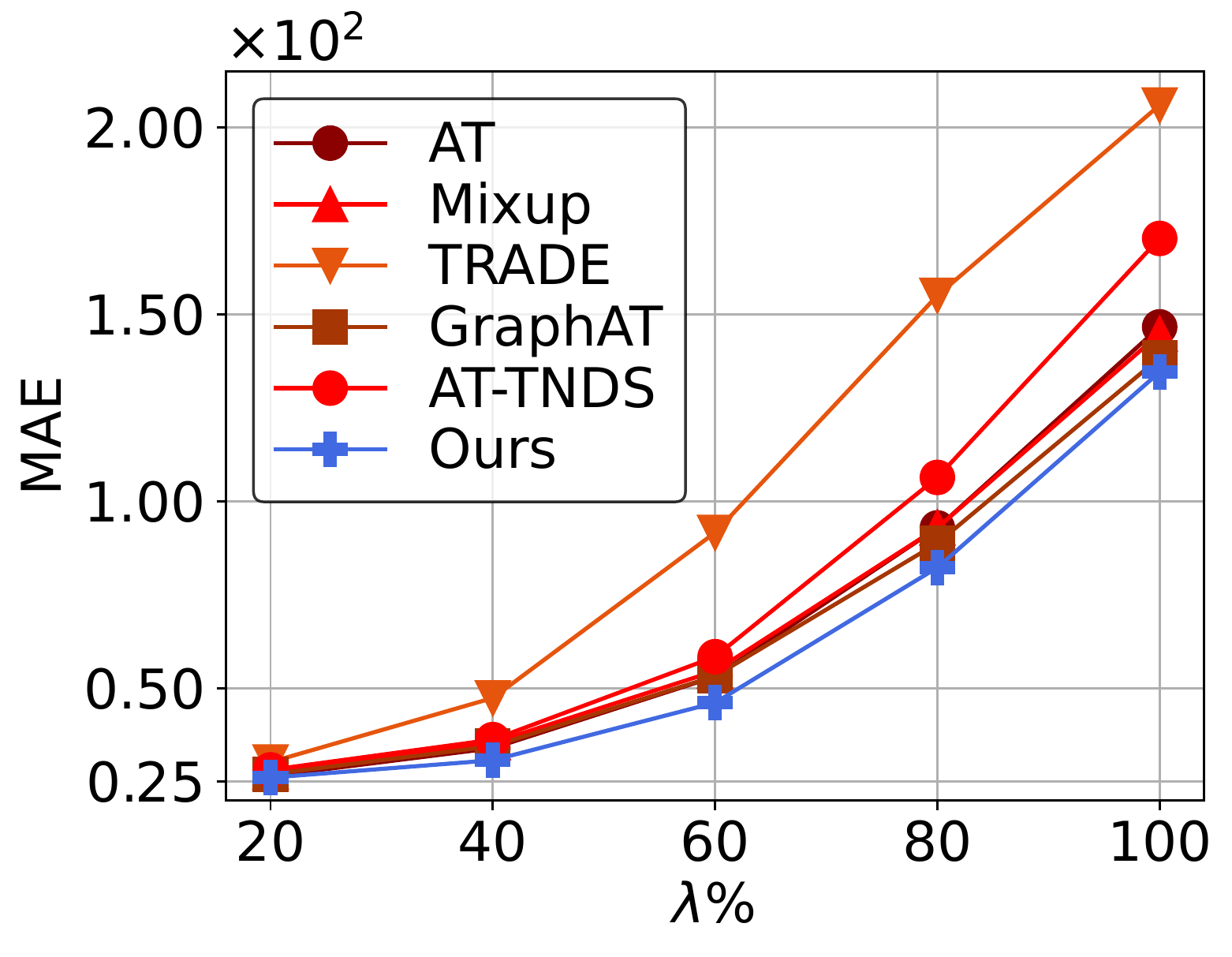}
\end{minipage}%
}%
\subfigure[PGD-Centrality]{
\begin{minipage}[]{0.25\linewidth}
\centering
\includegraphics[width=1.2in]{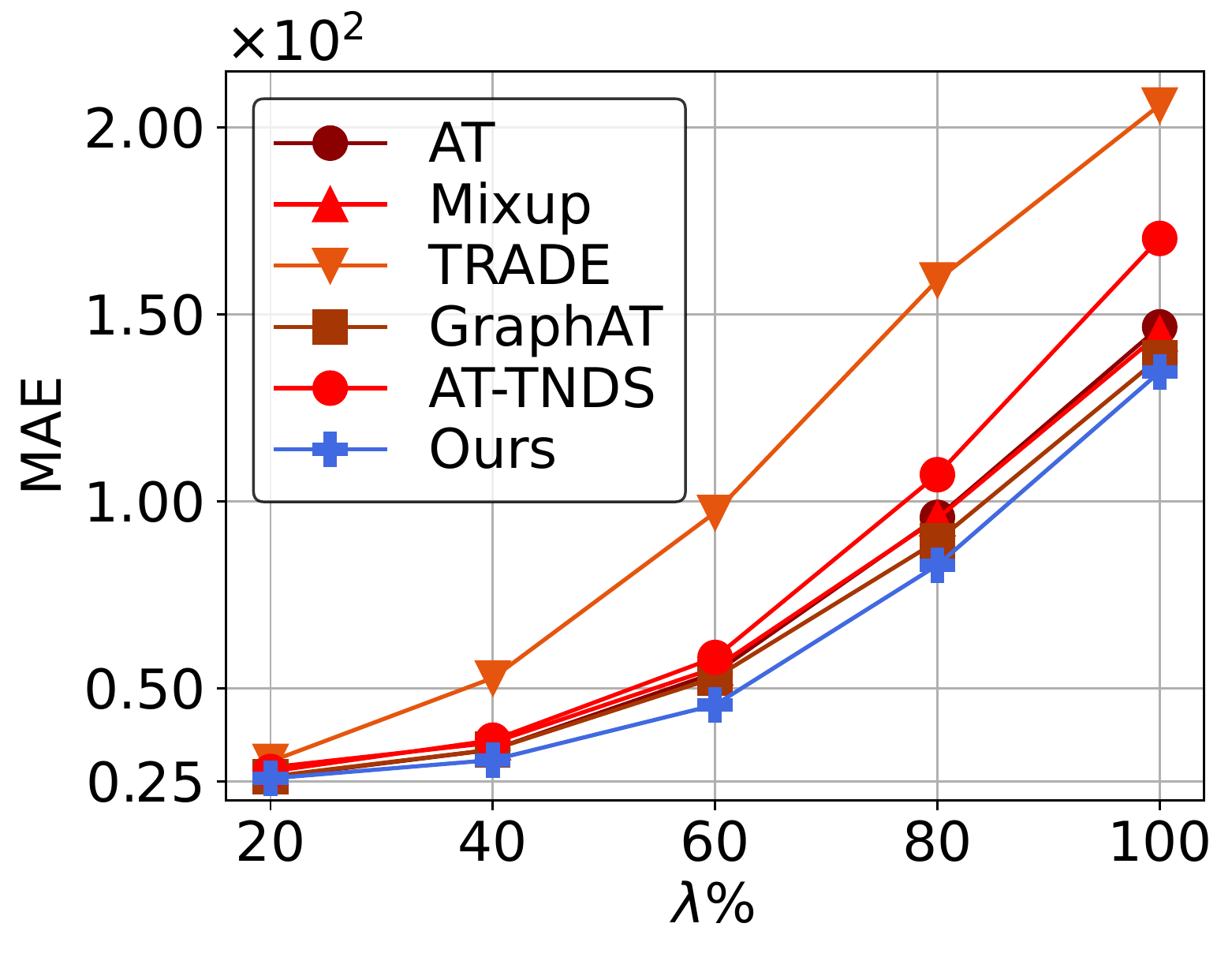}
\end{minipage}
}%
\subfigure[PGD-Degree]{
\begin{minipage}[]{0.25\linewidth}
\centering
\includegraphics[width=1.2in]{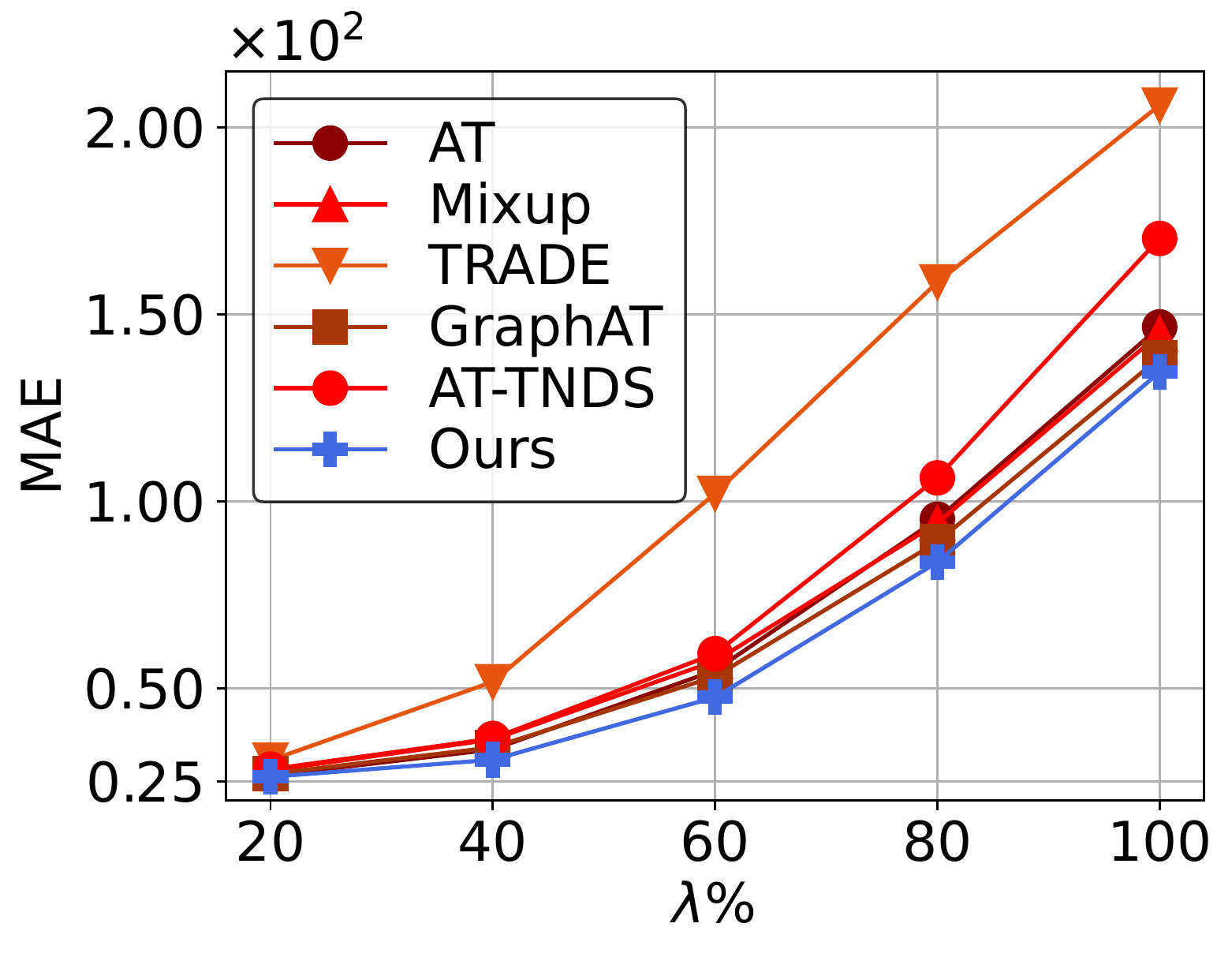}
\end{minipage}
}%
\subfigure[PGD-TNDS]{
\begin{minipage}[]{0.25\linewidth}
\centering
\includegraphics[width=1.2in]{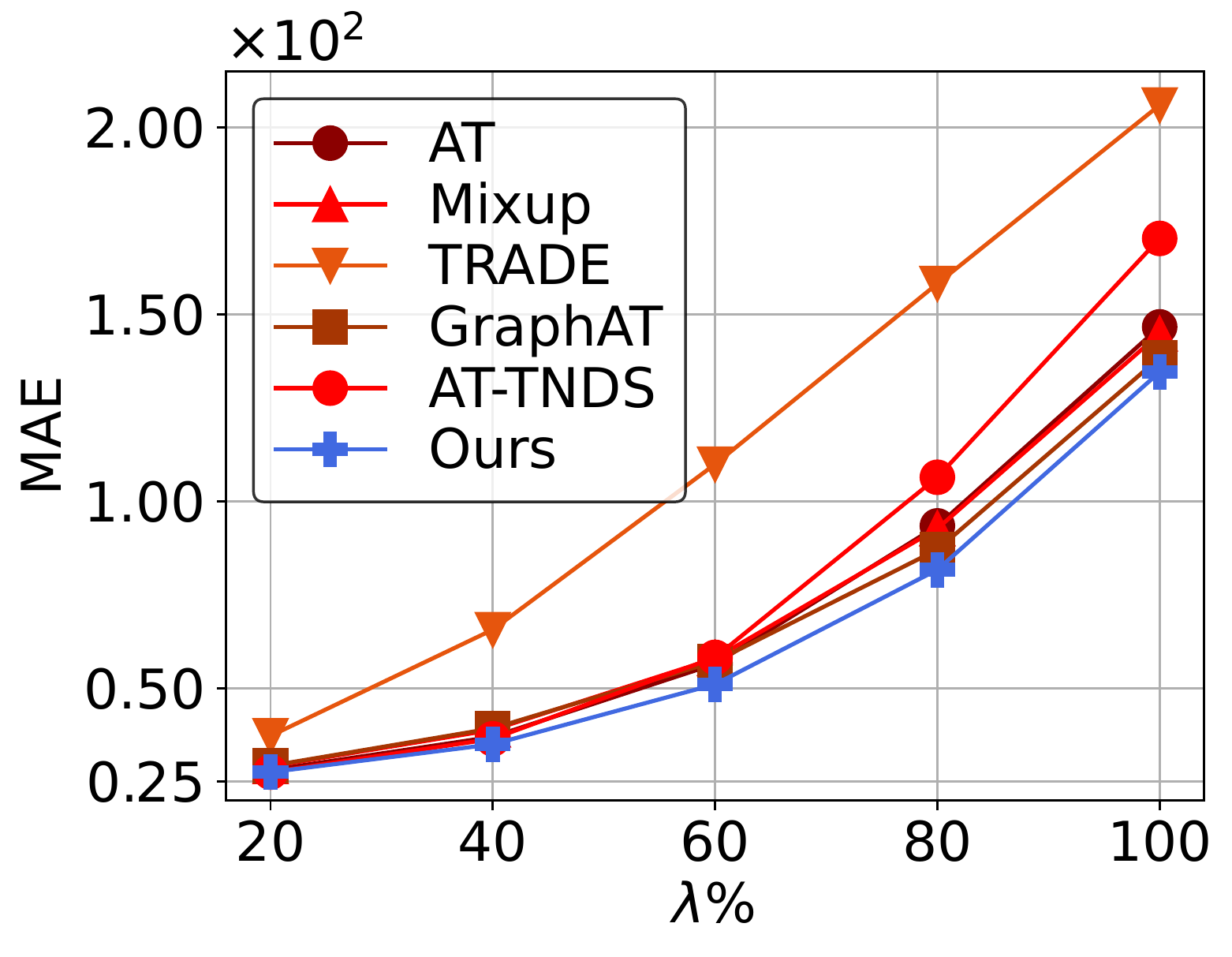}
\end{minipage}
}%
\centering
\caption{ Adversarial robustness performance under different attack strength on PeMS-D4}\label{fig:PeMS-D4-DAS}
\end{figure*}

\begin{figure*}[thb]
\centering
\subfigure[PGD-PR]{
\begin{minipage}[]{0.25\linewidth}
\centering
\includegraphics[width=1.2in]{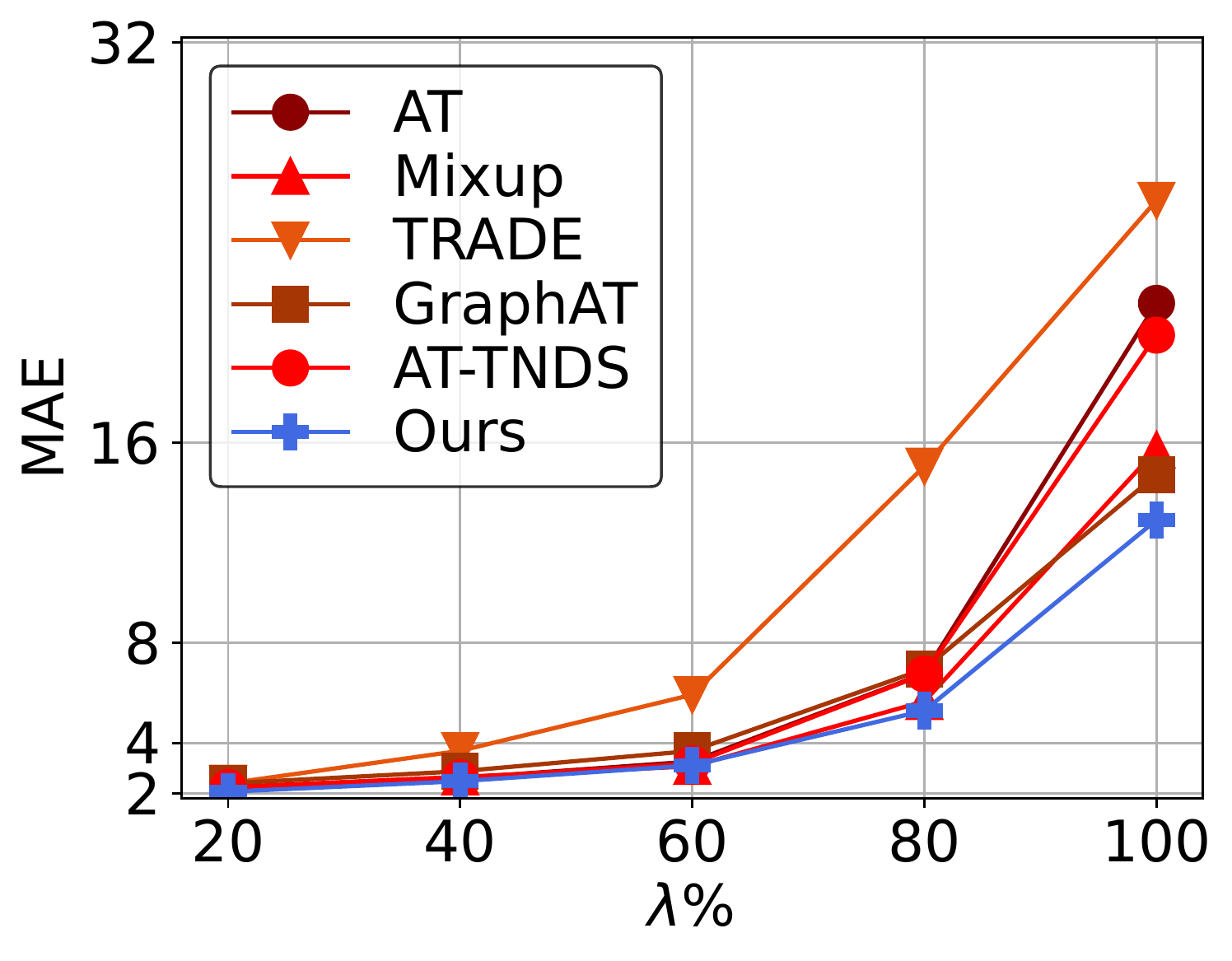}
\end{minipage}%
}%
\subfigure[PGD-Centrality]{
\begin{minipage}[]{0.25\linewidth}
\centering
\includegraphics[width=1.2in]{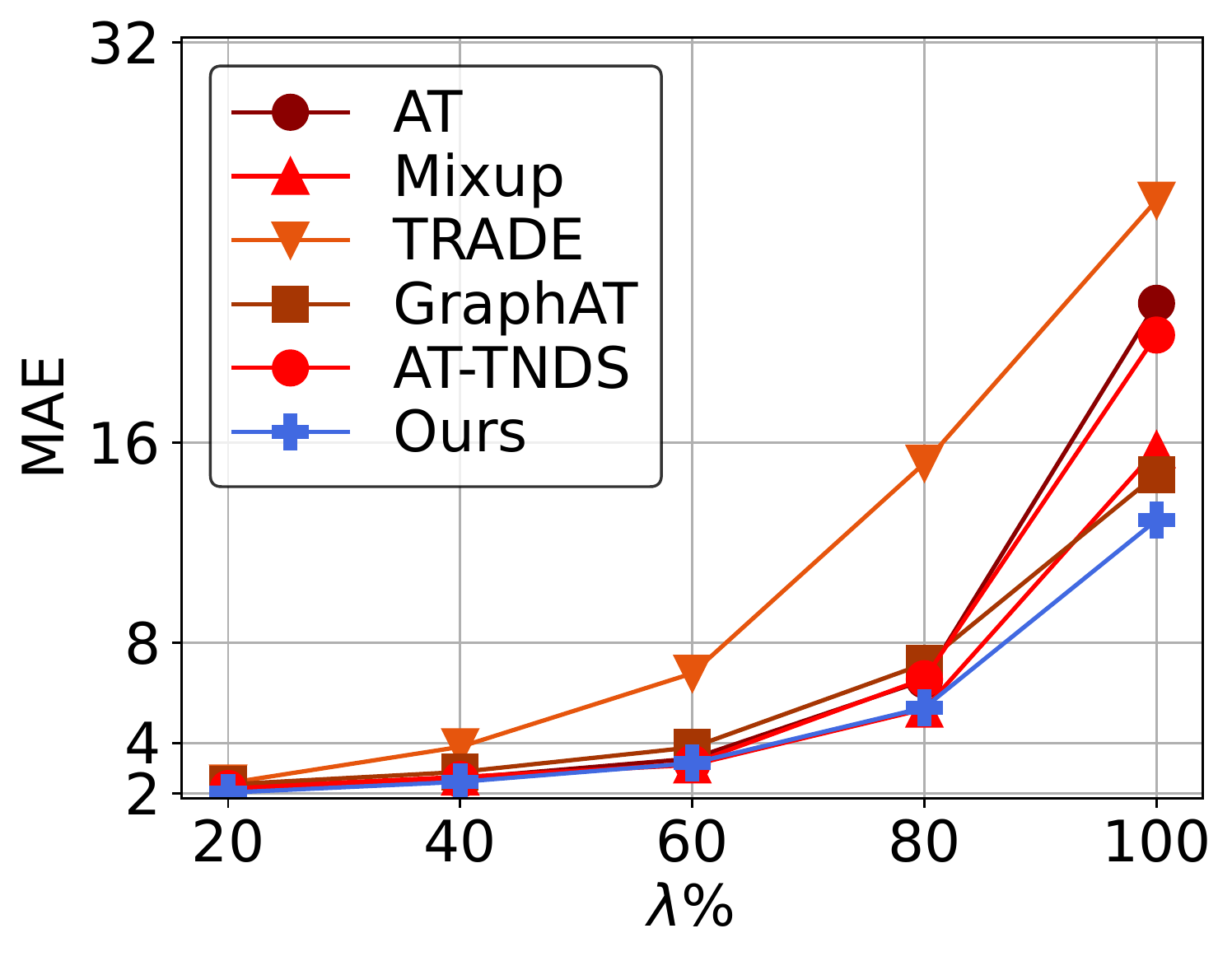}
\end{minipage}
}%
\subfigure[PGD-Degree]{
\begin{minipage}[]{0.25\linewidth}
\centering
\includegraphics[width=1.2in]{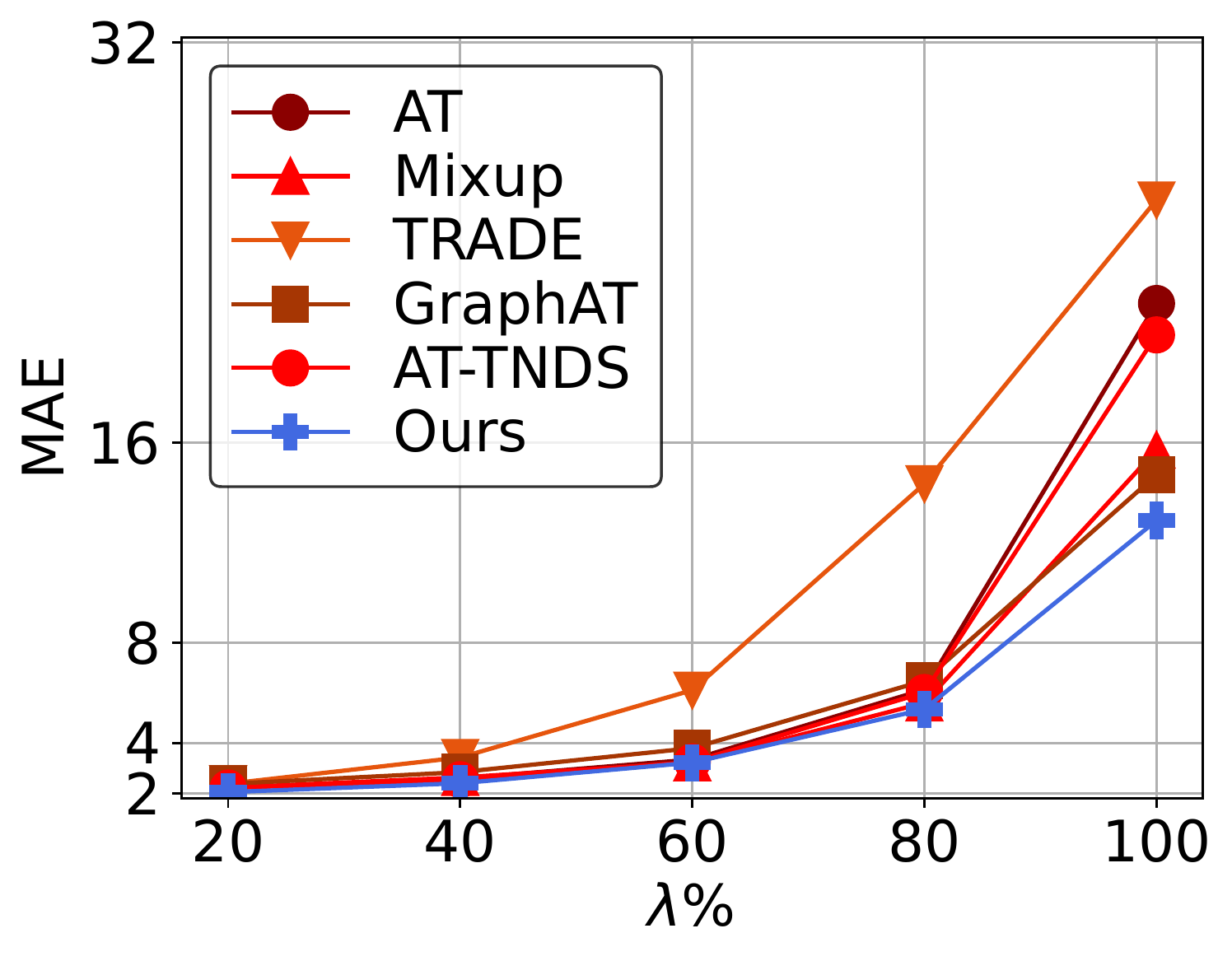}
\end{minipage}
}%
\subfigure[PGD-TNDS]{
\begin{minipage}[]{0.25\linewidth}
\centering
\includegraphics[width=1.2in]{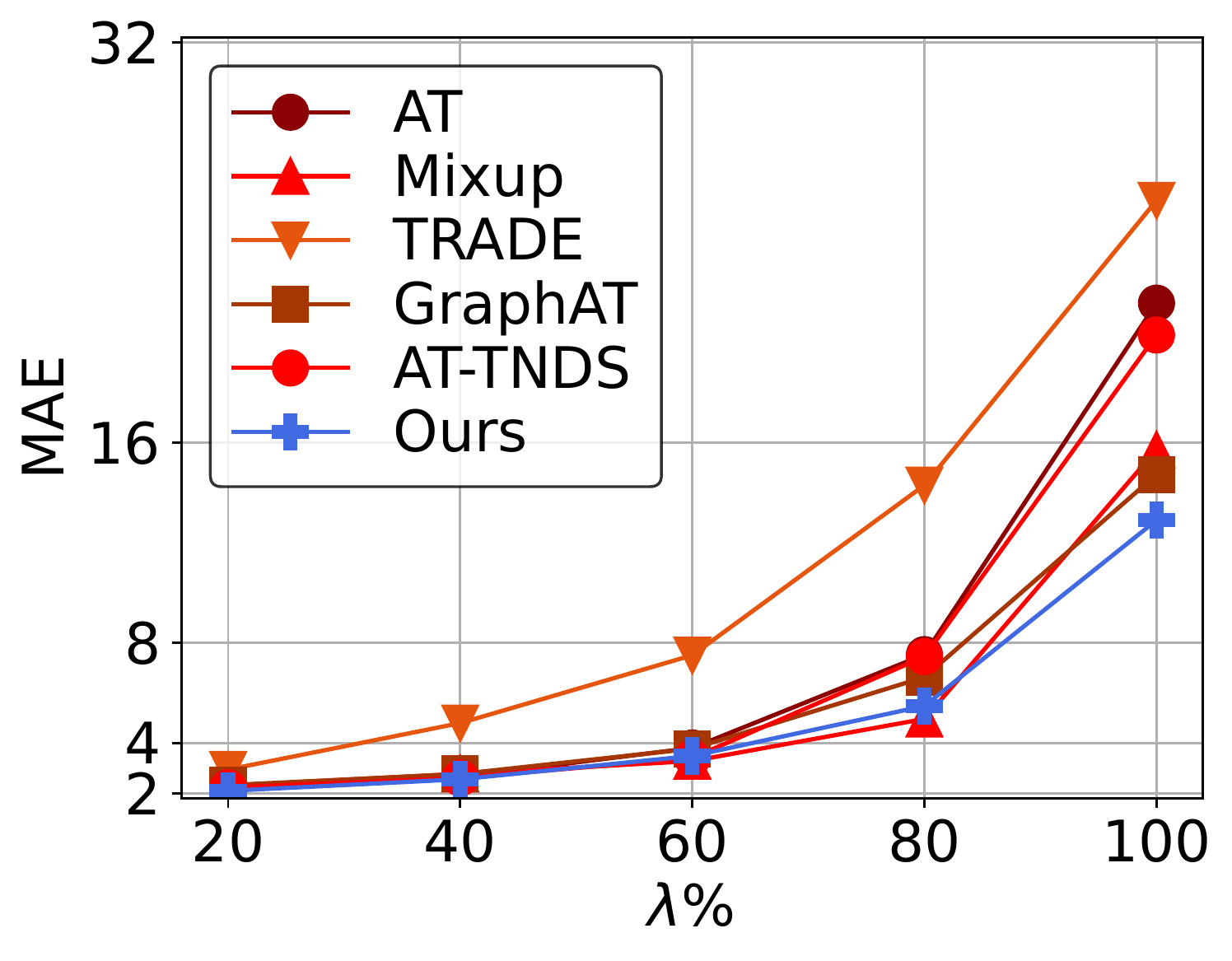}
\end{minipage}
}%
\centering
\caption{ Adversarial robustness performance under different attack strength on PeMS-BAY}\label{fig:PeMS-BAY-DAS}
\end{figure*}

The results of our adversarial robustness evaluation under four different attack strengths ($\lambda$=40,60,80,100), including PGD-PR, PGD-Centrality, PGD-Degree, and PGD-TNDS, on the PeMS-BAY and PeMS-D4 datasets are displayed in Figures \ref{fig:PeMS-D4-DAS} and \ref{fig:PeMS-BAY-DAS}. The results show that our method consistently outperforms other adversarial training methods in terms of adversarial robustness, regardless of the intensity of the attack.

\end{document}